\documentclass[11pt]{article}

\usepackage[preprint]{acl}

\usepackage{times}
\usepackage{latexsym}

\usepackage[T1]{fontenc}
\usepackage[utf8]{inputenc}

\usepackage{microtype}

\usepackage{inconsolata}

\usepackage{graphicx}

\usepackage{algorithm}      % 提供 \begin{algorithm} 浮动环境
\usepackage{algpseudocode}  % 提供 \State, \If, \For 等伪代码命令

\usepackage{booktabs}

\usepackage{amsmath}
\usepackage{amssymb}
\usepackage{mathtools}
\usepackage{amsthm}

\theoremstyle{definition}
\newtheorem{definition}{Definition}% 去掉 [section] 则全文连续编号
\usepackage[table]{xcolor}
\definecolor{graytext}{gray}{0.5}

\usepackage{multirow}

\usepackage{xcolor}
\usepackage[most]{tcolorbox} % 'most' 加载大多数常用库

\definecolor{prompt-gray}{RGB}{60, 60, 60}   % 标题栏深灰色
\definecolor{prompt-blue}{RGB}{65, 105, 225} % 底部高亮蓝色

\usepackage{listings}
\usepackage{enumitem}
\usepackage{tcolorbox}

\usepackage{tikz}
\usetikzlibrary{positioning, arrows.meta, shapes, calc}
\title{From Observation to Intervention: Distilling Self-Generated In-Context Demonstrations via Executable Causal Intervention}

\author{
 \textbf{Minming Ye\textsuperscript{1,2}}\quad
 \textbf{Min Liu\textsuperscript{2,\textdagger}}\quad
 \textbf{Haoang Chi\textsuperscript{3}}\quad
 \textbf{Ciheng Wu\textsuperscript{2}}\quad
 \textbf{Naiyang Guan\textsuperscript{2,\textdagger}}
\\
\\
 \textsuperscript{1}Shanghai Jiaotong University\quad
 \textsuperscript{2}Academy of Military Science
 \\
 \textsuperscript{3}National University of Defense Technology
 \\
 \small{\textdagger Corresponding author}
}

\begin{document}
\maketitle
\begin{abstract}
Self-generated In-Context Learning (ICL) enables LLMs to synthesize their own few-shot demonstrations, often outperforming manual prompts. However, existing methods filter demonstrations by observation, assuming a rationale is high-quality if it leads to the correct output. We show such filtering suffers from \textit{observational equivalence}, whereby both faithful causal rationales and spurious post-hoc rationales can yield same outputs and are statistically indistinguishable from observation alone. Inspired by independent causal mechanisms, we argue that faithful rationales imply \textit{transfer invariance}, in which causal rationales should remain effective across logically similar samples. We instantiate this criterion as \textbf{E}xecutable \textbf{C}ausal \textbf{I}ntervention for \textbf{D}emonstrations \textbf{D}istillation (ECI2D), which abstracts language rationales into executable programs and applies the causal interventions ($do$-operator) on logically similar samples, retaining only those demonstrations that imply transfer invariance. Experiments on math, logical and commonsense reasoning benchmarks demonstrate that ECI2D consistently outperforms observation-based ICL while retaining far fewer demonstrations, with gains exceeding 7 points as the test distribution shifts.
\end{abstract}

\section{Introduction}
\label{sec:intro}

\begin{figure}[ht]
  \centering
  \includegraphics[width=\linewidth]{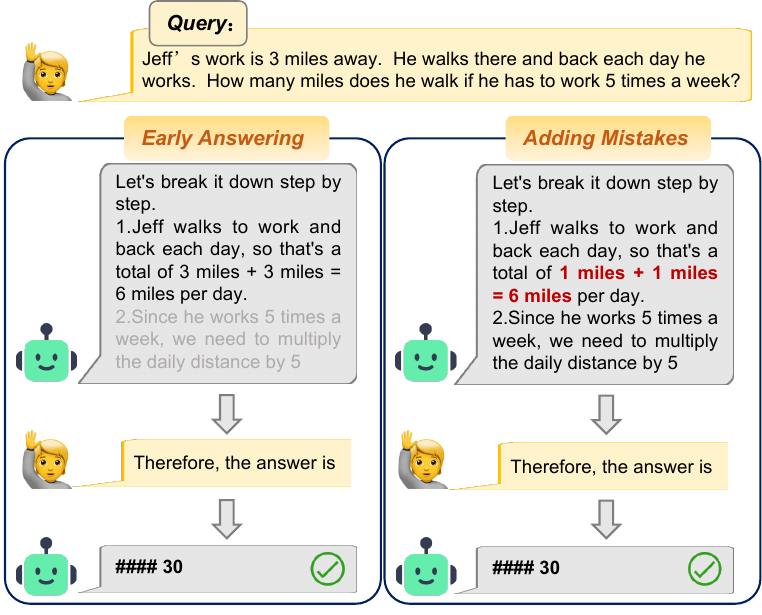}
  \caption{
    Illustration of post-hoc rationalization using \texttt{Llama3-8B} on the GSM8K~\cite{gsm8k}. \textbf{Early Answering} denotes the original CoT is truncated (denoted by \textcolor[RGB]{175, 171, 171}{gray}) to test if the answer is predetermined, and \textbf{Adding Mistakes} denotes a logical error is injected (highlighted in \textcolor[RGB]{192, 0, 0}{red}) before answering the question.
  }
  \label{fig:2}
\end{figure}

Few-shot in-context learning (ICL) with chain-of-thought (rationale) has become a cornerstone for eliciting reasoning in large language models (LLMs)~\cite{icl_survey,gpt3, cot,rethinking_icl,what_good_icl}. To reduce manual annotation, recent works focus on \textit{self-generated ICL}, which leverages the generative capacity of LLMs to synthesize in-context demonstrations~\cite{from_few_to_many,many_shot,auto_icl,are_human_gen,self_icl,self_gen_icl,auto_cot}.
Among them, reinforced ICL~\cite{many_shot,beyond_human_data} samples multiple rationales per training query and retains only those yielding the correct output, reporting gains over human annotation~\cite{are_human_gen,self_gen_icl}. We categorize them as \textbf{Self-Generated Reinforced ICL}.

\begin{figure}[t]
\centering
\begin{tikzpicture}[
    >={Latex[length=1.6mm,width=1.4mm]},
    every node/.style={font=\sffamily\footnotesize},
    var/.style    ={circle, draw, minimum size=0.78cm,
                    line width=0.7pt, inner sep=0pt},
    obs/.style    ={var, fill=blue!8,   draw=blue!50!black,  text=blue!25!black},
    causal/.style ={var, fill=green!12, draw=green!42!black, text=green!28!black},
    spurious/.style={var, fill=red!10,  draw=red!55!black,   text=red!40!black},
    neutral/.style={var, fill=gray!6,   draw=gray!55!black},
    oedge/.style  ={->, line width=0.85pt, draw=blue!50!black},
    cedge/.style  ={->, line width=0.95pt, draw=green!42!black},
    sedge/.style  ={->, line width=0.75pt, draw=red!55!black, dashed},
    sub/.style    ={font=\sffamily\scriptsize, text=gray!45!black},
    ptag/.style   ={font=\sffamily\small\bfseries, text=gray!40!black},
    banner/.style ={rectangle, rounded corners=2.5pt,
                    fill=gray!13, draw=gray!35, line width=0.35pt,
                    inner xsep=8pt, inner ysep=3.5pt,
                    font=\sffamily\scriptsize},
]

% ====== Panel (a): Observed SCM ======
\node[ptag] at (-3.5, 0) {(a)};
\node[obs] (Xa) at (-2.5, 0) {$X$};
\node[obs] (Ra) at ( 0.0, 0) {$R_{o}$};
\node[obs] (Ya) at ( 2.5, 0) {$Y$};
\draw[oedge] (Xa) -- (Ra);
\draw[oedge] (Ra) -- (Ya);
\node[sub, below=1pt of Xa] {input};
\node[sub, below=1pt of Ra] {observed};
\node[sub, below=1pt of Ya] {output};

% ====== Panel (b): True SCM (flat layout, no U) ======
\node[ptag] at (-3.5, -1.50) {(b)};

\node[neutral] (Xb) at (-2.5, -1.50) {$X$};
\node[causal]  (Rc) at ( 0.0, -1.50) {$R_{c}$};
\node[neutral] (Yb) at ( 2.5, -1.50) {$Y$};
\node[spurious](Rs) at ( 0.0, -2.55) {$R_{s}$};

% Faithful causal path
\draw[cedge] (Xb) -- (Rc);
\draw[cedge] (Rc) -- (Yb);

% Spurious collider
\draw[sedge] (Xb) -- (Rs);
\draw[sedge] (Yb) -- (Rs);

\node[sub, below=1pt of Xb] {input};
\node[sub, below=1pt of Yb] {output};

% ====== Bottom equivalence bar ======
\node[banner] at (0, -3.40)
    {\textit{Observational equivalence:}\, $P(Y\mid X,R)$ identical (Section~\ref{sec:problem_formulation})};

\end{tikzpicture}
\caption{%
Structural Causal Models (SCMs) in LLM inference. (a) \textbf{Observed SCM:} It treats the observed rationale as the direct causal mediator, forming a linear chain $X \to R_o \to Y$. (b) \textbf{True SCM:} The actual mechanism comprises a faithful causal path $X \to R_c \to Y$ and a spurious post-hoc path where $R_s$ is dependent on both $X$ and $Y$ ($X \dashrightarrow R_s \dashleftarrow Y$). The two SCMs are \textit{observational equivalence} as discussed in Section~\ref{sec:problem_formulation}.
}
\label{fig:3}
\end{figure}

However, the demonstration distillation criterion of self-generated reinforced ICL simply relies on \textit{observation}, which is a outcome-oriented criterion that equates answer correctness with rationale quality. Rationales are not always faithful, as shown in Figure~\ref{fig:2}, LLMs often engage in post-hoc rationalization~\cite{post-hoc}, generating plausible explanations after answering~\cite{llm_dont_say}. The faithfulness of post-hoc rationales cannot be guaranteed. We argue that observational distillation suffers from \textit{observational equivalence}. As shown in Figure~\ref{fig:3}, based on outcome correctness, the true inference process (comprising both faithful $X \to R_c \to Y$ and spurious $Y \dashrightarrow R_s$) is statistically indistinguishable from the observed causal chain $X \to R_o \to Y$. Self-generated reinforced ICL fails to disentangle $R_c$ from $R_s$, treating them as a single observed variable $R_o$, which leads to the problem of \textit{observational equivalence}. Critically, our experiments confirm that these unreliable rationales actively misguide the model (Table~\ref{tab:1}), degrading ICL performance. This naturally raises a question that needs attention:

\textit{How can we reliably distinguish faithful causal rationales from spurious post-hoc justifications?}

Inspired by the causal invariance~\cite{causal_invariant}, transportability~\cite{transportability}, and independent causal mechanisms~\cite{independent_causal_mechanisms}, we adopt a practical notion that faithful rationales should exhibit \textit{transfer invariance}, remaining effective across similar samples. To this end, we propose \textbf{E}xecutable  \textbf{C}ausal \textbf{I}ntervention for \textbf{D}emonstration \textbf{D}istillation (ECI2D), a novel framework to distill $R_c$ from $R_o$ via executable intervention. ECI2D first abstracts language rationales into executable programs. It then generates rationales and intervenes on logically similar samples by replacing the original rationale with these alternatives. We define an intervention-invariant scoring function to quantify transfer invariance, and only those rationales that consistently yield correct results across interventions are identified as the faithful causal path $R_c$, while those that fail are discarded as spurious $R_s$. Our contributions are fourfold:
\begin{itemize}
    \item We provide a formal causal analysis of self-generated reinforced ICL, identifying the prevalence of \textit{observational equivalence} and showing that faithful causal rationales imply \textit{transfer invariance} (Section~\ref{sec:problem_formulation}).
    \item We propose ECI2D, a framework that utilizes executable programs to perform causal interventions, effectively distilling faithful rationales (Section~\ref{sec:method}).
    \item We implement two abstraction forms, showing ECI2D can generalize from deterministic to non-deterministic tasks (Section~\ref{sec:method}).
    \item Extensive experiments demonstrate that ECI2D significantly improves performance and OOD generalization compared to both manual and self-generated reinforced ICL baselines (Section~\ref{sec:experimental_setup} and~\ref{sec:experimental_additional}).
\end{itemize}

\section{Related Works}
\label{sec:related_work}

\subsection{Self-Generated In-Context Learning}
Early works~\cite{self_gen_icl} synthesize demonstrations conditioned on test
inputs. Auto-ICL~\cite{auto_icl}, Self-ICL~\cite{self_icl}, and
AutoCoT~\cite{auto_cot} extend this into fully automated pipelines that bootstrap
demonstrations from zero-shot inference, surpassing human-crafted counterparts in
complex reasoning~\cite{are_human_gen}. Reinforced ICL~\cite{many_shot} instead
applies an outcome-oriented criterion, retaining only the rationale traces that
yield the correct final answer.

This criterion cannot guarantee that a retained rationale reflects the process
producing the answer, raising the concern that unfaithful reasoning enters the
prompt and degrades generalization~\cite{llm_dont_say, measuring_cot,
towards_understanding}. Prior studies diagnose this through observational analysis
or input perturbations~\cite{towards_faithfully_nlp, cot}. We instead propose a novel framework that utilizes interventional transfer invariance to rigorously quantify and verify the faithfulness of rationales.

\subsection{Causal Abstraction \& Externally Verified Reasoning}
Causal abstraction interprets LLM behaviour by mapping natural language to
structured causal models~\cite{neural_natural_language, wu2023interpretability,
yamakoshi2023causal}. Separately, program-aided prompting executes generated
code~\cite{pal, pot, coc} and symbolic approaches delegate a logical form to a
solver~\cite{logiclm, satlm, linc}, both requiring an external engine at inference
time. We combine the two by applying causal abstraction to the rationale itself,
whose logical steps correspond to causal nodes~\cite{causal_abstraction}, and
converting it into an executable program that enables precise intervention during
demonstration distillation rather than inference.

\section{Problem Analysis}
\label{sec:problem_formulation}
% 以该新章节为准，讲故事，not堆砌概念，正文给结论，附录给分析
The objective of self-generated reinforced ICL is to construct a high-quality demonstration set $D$ from a training set defined over input space $\mathcal{X}$, output space $\mathcal{Y}$, and rationale space $\mathcal{R}$. Unlike manual curation, the LLM with parameters $\Theta$ autonomously synthesizes demonstrations in two phases. The Generation phase samples a candidate pair $(\hat{r}, \hat{y}) \sim \Theta(\cdot \mid x)$ given $x \in \mathcal{X}$, where $\hat{r} \in \mathcal{R}$ and $\hat{y} \in \mathcal{Y}$. The Distillation phase filters candidates using a scoring function $S(x, \hat{r}, \hat{y})$. Existing methods predominantly adopt $S(x, \hat{r}, \hat{y}) = \mathbb{I}(\hat{y} = y)$, relying solely on observational correctness relative to ground truth $y$.

\begin{figure*}[ht]
  \centering 
  \includegraphics[width=\linewidth]{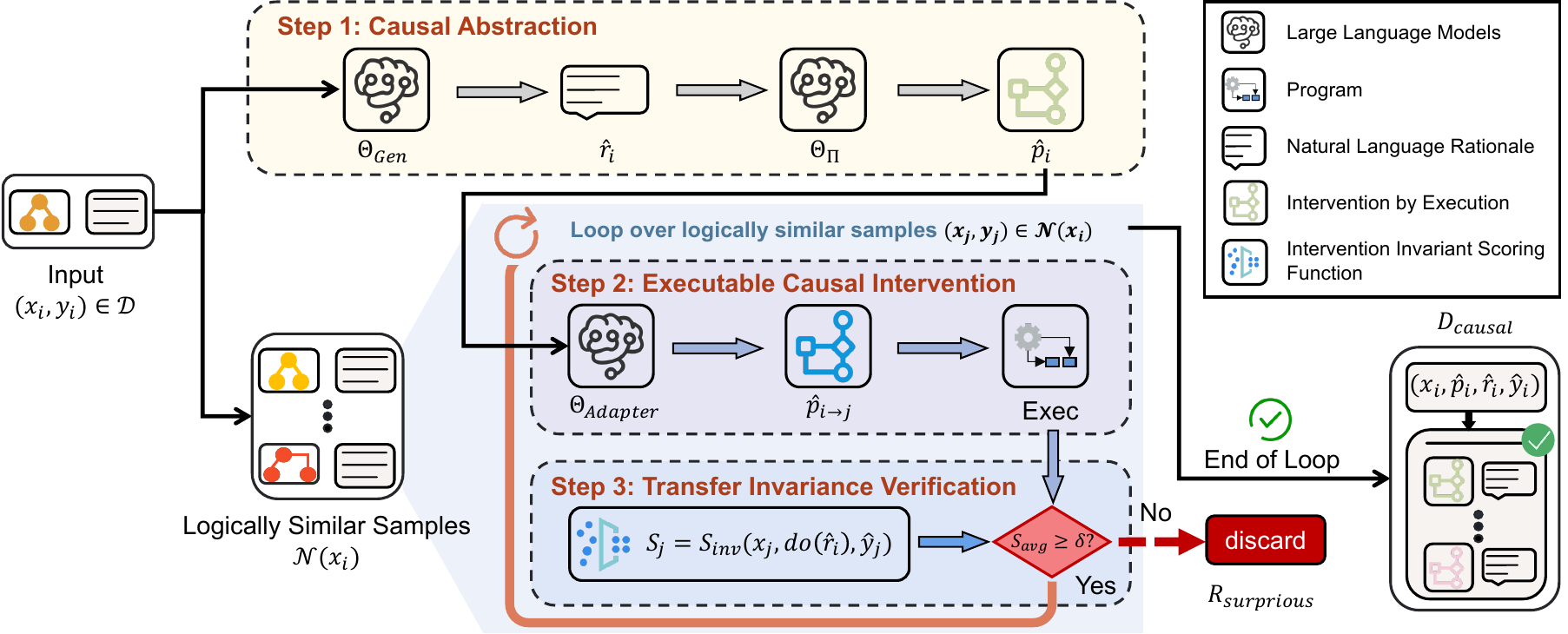}
    \caption{
      Invariant Reasoning Distillation phase in ECI2D. ECI2D distills faithful causal demonstrations through an \textbf{Abstract-Intervene-Verify} pipeline. \textit{Step 1}: Causal Abstraction. For each sample $(x_i, y_i)$, the LLM generator $\Theta_{\text{Gen}}$ produces a natural language rationale $\hat{r}_i$, which is then abstracted by LLM Abstractor $\Theta_{\Pi}$ into an executable program $\hat{p}_i$. \textit{Step 2}: Executable Causal Intervention. For all logically similar samples $(x_j, y_j) \in \mathcal{N}(x_i)$, LLM adapter $\Theta_{\text{Adapter}}$ transplants the logical skeleton of $\hat{p}_i$ to the logically similar inputs to generate $\hat{p}_{i \to j}$, performing an executable intervention $do(R=\hat{r}_i)$ on $(x_j,y_j)$ via execution. \textit{Step 3}: Transfer Invariance Verification. The intervention output is scored against the ground truth $y_j$. Rationales that fail to generalize are identified as spurious and discarded, while those demonstrating transfer invariance are retained in the causal pool $D_{\text{causal}}$.
    }
  \label{fig:4}
\end{figure*}

\subsection{The Limitation of Self-Generated Reinforced ICL}
The criterion of observational distillation assumes that answer correctness guarantees rationale quality. However, it overlooks post-hoc rationalization in LLM inference. LLMs may determine answers via pattern matching or memorized shortcuts, then retroactively generate plausible-sounding rationales to justify the conclusion~\cite{measuring_cot,post-hoc}. Since these rationales are downstream of the prediction, they lack genuine influence on output.

To further analyze this limitation, we model the inference process using SCM $\mathcal{G} = \langle \mathbf{V}, \mathbf{E} \rangle$~\cite{scm}, where $\mathbf{V} = \{X, R, Y\}$ represents the input, rationale, and output, respectively. As shown in Figure~\ref{fig:3}, current distillation methods implicitly assume an observed causal graph $X \to R_o \to Y$, treating the rationale as a direct mediator of $Y$. We identify the fundamental limitation of existing distillation methods as the inability to differentiate between faithful causal $R_c$ and spurious post-hoc rationales $R_s$ based solely on observation, which suffers from \textit{observational equivalence}.

\begin{definition}[Faithful Causal Rationale]
\label{def:faithful}
A faithful causal rationale is one that mediates the inference and has a genuine causal effect on the output. 
\end{definition}

\begin{definition}[Spurious Post-hoc Rationale]
\label{def:spurious}
A post-hoc rationale is an explanation that is generated after the prediction has been determined. A post-hoc rationale is considered spurious if it is retrospectively constructed to fit the observed input and output, while failing to reflect the causal factors that actually produced the prediction.
\end{definition}

\begin{definition}[Observational Equivalence]
\label{def:equivalence}
Two causal structures are observationally equivalent if they generate identical joint probability distributions over the observed variables. 
\end{definition}

As derived in Appendix~\ref{app:derivation}, given a demonstration $(x, \hat{r}, y)$, both the faithful and the spurious factorization produce identical observational likelihoods $P(Y=y \mid X=x, R=\hat{r})$. Consequently, outcome-oriented  filtering suffers from \textit{observational equivalence}, rendering the true causal structure unidentifiable~\cite{verma2022equivalence,controlling_selection,spirtes2000causation}. 

\subsection{Transfer Invariance as a Proxy for Faithfulness}

To resolve the \textit{observational equivalence}, it is imperative to distinguish between the faithful causal rationale $R_c$ and the spurious post-hoc rationale $R_s$. As analyzed in Appendix~\ref{app:icm_proof}, inspired by the Independent Causal Mechanisms~\cite{independent_causal_mechanisms}, we adopt the practical notion that a faithful causal relation should remain effective across similar samples~\cite{causal_invariant,transportability}, which implies \textit{transfer invariance}.

Consider a faithful rationale $\hat{r}_i \in \mathcal{R}$ derived from an input $x_i \in \mathcal{X}$. If $\hat{r}_i$ represents the true causal rationale of output $y_i$, it should be applicable to a logically similar sample $x_j \in \mathcal{N}(x_i)$ without losing its validity, where $\mathcal{N}(x_i)$ denotes the logically similar sample space of $x_i$. Formally, Sample $Q$ can be characterized by logical features $A=(P, S, T)$, where $P$, $S$, and $T$ represent premises, subjects, and result type, respectively. Discrete structural distance is defined as $\theta(A_i, A_j) \in \{0, 1\}$. Two samples $Q_i$ and $Q_j$ are logically similar if $\theta(A_i, A_j)=0$. The logically similar samples are thus $\mathcal{N}(x_i) = \{x_j \mid \theta(A_i, A_j) = 0\}$. Faithfulness implies that applying $\hat{r}_i$ to $\mathcal{N}(x_i)$ preserves predictive correctness under $do$ intervention:
\begin{equation}
\label{eq:intervention}
\begin{split}
    &P(Y=y_i \mid do(R=\hat{r}_i), X=x_i) \\
    &\approx P(Y=y_j \mid do(R=\hat{r}_i), X=x_j).
\end{split}
\end{equation}
This formulation rests on two complementary grounds. Faithfulness implies invariance, since any rationale that fails to remain valid across logically similar samples cannot be the underlying causal mechanism that mediates the inference. A rationale also derives from being a procedure connecting inputs to outputs, so a rationale that fails to transfer beyond its source input may be a post-hoc label rather than a causal explanation. Equation~\ref{eq:intervention} shifts distillation from passive observation to active intervention, directly motivating ECI2D (Section~\ref{sec:method}).

Additionally, the inference of LLMs is inevitably susceptible to latent confounders~\cite{chi2024unveiling}. However, transfer invariance circumvents most latent confounders, such as shortcuts in LLM memory, which fail to satisfy Equation~\ref{eq:intervention} due to their limited generalizability across similar samples.

\section{The ECI2D framework}
\label{sec:method}
ECI2D shifts the paradigm from observation to intervention and consists of two phases: (1) Invariant Reasoning Distillation, where we construct a pool of causal rationales by verifying transfer invariance, and (2) Causal Steering Inference, where validated demonstrations guide the model during inference.

As shown in Figure~\ref{fig:4} and Algorithm~\ref{alg:ECI2D_distillation}, during the Invariant Reasoning Distillation phase, we introduce an
\textbf{Abstract-Intervene-Verify} pipeline whose abstraction step maps a
natural language rationale to a program, implemented as a Python
program in the main experiments
(Section~\ref{sec:invariant_reasoning_distillation}) and as a hybrid program that
extends the intervention principle to semantic reasoning (Section~\ref{sec:instantiation}).

\subsection{Invariant Reasoning Distillation with Python Program}
\label{sec:invariant_reasoning_distillation}

\paragraph{Step 1: Causal Abstraction.}
Previous work has argued that rationales can be interpreted as implicit causal models~\cite{causal_abstraction}; however, rationales expressed in natural language are high-dimensional, ambiguous, and entangled with specific contexts. To enable precise intervention, we convert these rationales into deterministic Python programs~\cite{coc,pal,pot} via causal abstraction mapping function $\Pi: \mathcal{R} \to \mathcal{P}$. 

For training sample $(x_i, y_i)$, an LLM Generator $\Theta_{\text{Gen}}$ generates a natural language rationale $\hat{r}_i$:
\begin{equation}
\label{eq:gen}
    \hat{r}_i, \hat{y}_i = \Theta_{\text{Gen}}(x_i).
\end{equation}
Then an LLM Abstractor $\Theta_\Pi$ is utilized to convert rationale $\hat{r}_i$ into a Python program $\hat{p}_i \in \mathcal{P}$:
\begin{equation}
\label{eq:abstract}
    \hat{p}_i = \Theta_\Pi(\hat{r}_i).
\end{equation}
Following self-generated reinforced ICL, we initially filter rationales using the outcome-oriented criterion. The program $\hat{p}_i$ serves as a causal abstraction of rationale $\hat{r}_i$, which allows us to perform precise causal interventions. In our experiment, generation and abstraction were implemented in a single LLM prompt shown in Appendix~\ref{app:prompt_gen_abs}.

\paragraph{Step 2: Executable Causal Intervention.}
Next, we utilize the executable program $\hat{p}_i$ to perform the causal intervention. However, since $x_i$ and $x_j$ contain different parameters, it is infeasible to directly execute $\hat{p}_i$ on $x_j$. To this end, we decompose program $\hat{p}_i$ into a tuple $\langle \mathrm{s}_i, \mathrm{v}_i \rangle$, where $\mathrm{s}_i$ denotes the logical skeleton, capturing the reasoning structure independent of specific parameters $\mathrm{v}_i$ from input $x_i$. We prompt an LLM Adapter to generate an adapted program $\hat{p}_{i\to j}$ by modifying only the parameters $\mathrm{v}_i$, while preserving the skeleton $\mathrm{s}_i$:
\begin{equation}
\label{eq:adapted}
    \hat{p}_{i\to j} = \Theta_\text{Adapter}(\hat{p}_i, x_j) = \langle \mathrm{s}_i, \mathrm{v}_j \rangle.
\end{equation}
Consequently, we can apply the intervention $do(R=\hat{r}_i)$ to $(x_j,y_j)$:
\begin{equation}
\label{eq:y_pred}
    \hat{y}_j=\text{Exec}(\hat{p}_{i \to j}),
\end{equation}
where $\hat{y}_j$ is the output after applying intervention $do(R=\hat{r}_i)$ to the logically similar sample $x_j$. The detailed prompts used to adapt and intervene are in Appendix~\ref{app:prompt_adapt}. Crucially, intervention targets are drawn from the logically similar samples $\mathcal{N}(x_i)$. We construct $\mathcal{N}(x_i)$ via a two-stage retrieval: (1) coarse semantic filtering using Sentence-Transformer\footnote{https://huggingface.co/sentence-transformers/all-MiniLM-L6-v2} and (2) fine-grained structural verification via an LLM that outputs a binary indicator $\theta \in \{0,1\}$ to judge equivalence of $A=(P, S, T)$. Detailed prompts are in Appendix~\ref{app:prompt_retriveal}.

\begin{table*}[t]
\centering
\caption{Performance comparison on math and logical reasoning benchmarks. Following standard CoT~\cite{cot}, we use 8-shot demonstrations for math and 3-shot for logical tasks. We report accuracy (\%) for all benchmarks except DROP (F1 \%). \textit{+ Code} denotes the self-generated reinforced ICL with program-augmented demonstrations, which is the
same rationale-and-program format as ECI2D. Retention rates after distillation are included. Best results per benchmark are \textbf{bold}, second-best are \underline{underlined}. GSM-S = GSM-Symbolic, P1/P2 = Plus-1/Plus-2, Date U. = Date Understanding, Obj. C. = Object Counting, Word S. = Word Sorting, Ret. = Retention}
\label{tab:1}
\small
\renewcommand{\arraystretch}{1.15}
\setlength{\tabcolsep}{3.5pt}
\begin{tabular}{ l *{6}{c} c *{5}{c} }
\toprule
\multirow{2}{*}{Method} & \multicolumn{6}{c}{\textbf{Math Reasoning}} & & \multicolumn{5}{c}{\textbf{Logical Reasoning}} \\
\cmidrule(lr){2-7} \cmidrule(lr){9-13}
 & GSM8K & SVAMP & GSM-S & P1 & P2 & Ret. & & DROP & Date U. & Obj. C. & Word S. & Ret. \\
\midrule

\rowcolor{gray!10}\multicolumn{13}{c}{\texttt{Meta-Llama-3-8B-Instruct}} \\
\midrule
Direct & 6.97 & 63.50 & 70.12 & 55.46 & 31.96 & -- & & 62.69 & 42.00 & 40.40 & 53.60 & -- \\
Zero-shot CoT & 71.95 & 80.10 & 67.16 & 54.06 & 29.76 & -- & & 74.94 & \underline{70.80} & 73.20 & 52.80 & -- \\
Manual Few-shot CoT & 79.00 & 81.30 & 75.32 & 58.54 & 32.04 & -- & & 76.47 & \textbf{73.20} & \textbf{86.00} & 54.00 & -- \\
Self-Gen. Reinforced ICL & \underline{81.88} & 84.80 & \underline{76.88} & \underline{63.74} & \underline{36.92} & 86\% & & 74.99 & 53.60 & 77.60 & 50.80 & 63\% \\
\quad \textit{+ Code} & 73.41 & \underline{85.10} & 74.92 & 62.16 & 36.84 & 85\% & & \underline{78.01} & 57.60 & 73.60 & \underline{55.20} & 61\% \\
\rowcolor{blue!8} ECI2D (ours) & \textbf{82.11} & \textbf{86.60} & \textbf{77.36} & \textbf{66.02} & \textbf{37.72} & 52\% & & \textbf{80.87} & 62.80 & \underline{80.00} & \textbf{59.60} & 27\% \\
\midrule

\rowcolor{gray!10}\multicolumn{13}{c}{\texttt{Qwen2.5-7B-Instruct}} \\
\midrule
Direct & 18.65 & 68.30 & 76.40 & 67.68 & 52.60 & -- & & 64.29 & 54.40 & 44.40 & 20.00 & -- \\
Zero-shot CoT & 81.88 & 81.20 & 76.80 & 67.94 & 56.16 & -- & & 74.68 & 75.20 & 76.80 & 22.00 & -- \\
Manual Few-shot CoT & 87.04 & 93.40 & 81.90 & 71.76 & 58.24 & -- & & 79.19 & \textbf{79.20} & 77.20 & \textbf{29.20} & -- \\
Self-Gen. Reinforced ICL & \underline{91.58} & \underline{93.90} & \underline{86.54} & 74.94 & 58.12 & 93\% & & 72.73 & 75.60 & 79.60 & 22.00 & 71\% \\
\quad \textit{+ Code} & 84.93 & 92.00 & 86.24 & \underline{76.80} & \underline{62.10} & 90\% & & \underline{79.24} & 72.80 & \underline{83.60} & 23.20 & 70\% \\
\rowcolor{blue!8} ECI2D (ours) & \textbf{91.74} & \textbf{95.50} & \textbf{89.80} & \textbf{82.02} & \textbf{65.28} & 57\% & & \textbf{80.54} & \underline{78.00} & \textbf{85.20} & \underline{24.40} & 32\% \\
\bottomrule
\end{tabular}
\end{table*}

\paragraph{Step 3: Verification of Transfer Invariance.}
As discussed in Section~\ref{sec:problem_formulation}, a faithful rationale is expected to generalize across $\mathcal{N}(x_i)$. Since $P(Y=y_j \mid do(R=\hat{r}_i), X=x_j)$ in Equation~\ref{eq:intervention} cannot be computed in closed form for an LLM, we estimate it through Monte Carlo sampling over $\mathcal{N}(x_i)$. Each binary outcome $\mathbb{I}(\hat{y}_j = y_j)$ is an unbiased single-sample estimator 
of this probability at neighbor $x_j$, and their average over $N$ neighbors estimates how reliably $\hat{r}_i$ preserves correctness across $\mathcal{N}(x_i)$. The continuous tolerance $\approx$ in Equation~\ref{eq:intervention} is realized as a discrete threshold $\delta$ that determines whether $\hat{r}_i$ is retained. Formally, we define an intervention-invariant scoring function $S_{\text{inv}}$ that evaluates whether a rationale remains valid under causal intervention across related inputs:
\begin{equation}
S_{\text{inv}}(x_j, do(\hat{r}_i), \hat{y}_j) 
= \mathbb{I}\!\left( \hat{y}_j = y_j \right),
\end{equation}
where $\hat{y}_j$ is calculated via Equation~\ref{eq:y_pred} and $\mathbb{I}(\cdot)$ denotes the indicator function. For a sample $(x_i,y_i)$ and a set of logically similar samples $\{x_1,...,x_N\}\subset \mathcal{N}(x_i)$, the set of invariant causal demonstrations $D_{\text{causal}}$ is constructed as:
\begin{equation}
\left\{
(x_i,\hat{p}_i, \hat{r}_i,\hat{y}_i)\;\middle|\;
\frac{1}{N} \sum_{j=1}^{N} 
S_{\text{inv}}(x_j, do(\hat{r}_i), \hat{y}_j) \geq \delta 
\right\},
\end{equation}
where $(x_j,y_j)$ denotes the $j$-th logically similar sample from $\mathcal{N}(x_i)$ and $\delta$ denotes the threshold. Consequently, we explicitly store the tuple $(x_i,\hat{p}_i, \hat{r}_i,\hat{y}_i)$ to preserve both the executable program $\hat{p}_i$ and natural language rationale $\hat{r}_i$.

\subsection{Invariant Reasoning Distillation with Hybrid Program}
\label{sec:instantiation}
A Python program cannot express steps that require world knowledge, so the LLM
Abstractor $\Theta_\Pi$ of Section~\ref{sec:invariant_reasoning_distillation} cannot
abstract such rationales. Following Chain-of-Code~\cite{coc}, we widen ECI2D to
hybrid programs, in which each such step is mapped to a function declaration
answered by an LLM call. This requires generalizing only $\Theta_\Pi$ in Step 1 and
the intervention in Step 2.

\paragraph{Hybrid Causal Abstraction of $\Theta_\Pi$.}
In Equation~\ref{eq:abstract}, $\Theta_\Pi$ maps $\hat{r}_i$ into three parts,
\begin{equation}
\label{eq:hybrid_abstract}
    \hat{p}_i = \Theta_\Pi(\hat{r}_i) = \langle \mathrm{d}_i, \mathrm{s}_i, \mathrm{v}_i \rangle,
\end{equation}
where $\mathrm{d}_i$ are the declarations of the steps requiring world knowledge, each
carrying a type signature and a docstring, $\mathrm{s}_i$ is the Python logical
skeleton that calls them, and $\mathrm{v}_i$ are the parameters of $\mathrm{s}_i$.
Both $\mathrm{d}_i$ and $\mathrm{s}_i$ carry the reasoning structure, so the LLM
Adapter in Equation~\ref{eq:adapted} still modifies $\mathrm{v}_i$ alone to obtain
$\hat{p}_{i \to j} = \Theta_{\text{Adapter}}(\hat{p}_i, x_j) = \langle \mathrm{d}_i,
\mathrm{s}_i, \mathrm{v}_j \rangle$. Detailed prompts are in
Appendix~\ref{app:prompt_hybrid_abstraction}.

\paragraph{Intervention for Hybrid Program.}
Since $\mathrm{d}_i$ carries no function body, $\hat{p}_{i \to j}$ is not executable.
Each declaration is therefore resolved by an LLM Evaluator $\Theta_{\text{Eval}}$ at
the point where $\mathrm{s}_i$ calls it, which generalises
Equation~\ref{eq:y_pred} to
\begin{equation}
\label{eq:eval}
    \hat{y}_j = \mathrm{Exec}\big(\langle \Theta_{\text{Eval}}(\mathrm{d}_i,
    \mathrm{a}_j), \mathrm{s}_i, \mathrm{v}_j \rangle\big),
\end{equation}
where $\mathrm{a}_j$ denotes the arguments $\mathrm{s}_i$ passes at that call site.
$\Theta_{\text{Eval}}$ is given $\mathrm{d}_i$ and $\mathrm{a}_j$ alone, and its
answer is returned to $\mathrm{s}_i$, which resumes execution and produces
$\hat{y}_j$. The prompt for $\Theta_{\text{Eval}}$ is given in
Appendix~\ref{app:prompt_fun_eval}. The intervention $do(R = \hat{r}_i)$ on
$(x_j, y_j)$, the retrieval of logically similar samples, and the comparison of
$S_{\text{inv}}$ against $\delta$ are the same as Section~\ref{sec:invariant_reasoning_distillation}. Detailed prompts are in
Appendix~\ref{app:prompt_hybrid_intervention}.

\subsection{Causal Steering Inference}
During the inference phase, we leverage the distilled invariant causal demonstrations $D_{\text{causal}}$ to steer the reasoning process for test sets. Given a test query $x_{\text{test}}$, we retrieve the top-$k$ demonstrations using Sentence-Transformer from $D_{\text{causal}}$ to construct the few-shot ICL. Crucially, each retrieved demonstration consists of a tuple $(x,\hat{p},\hat{r},y)$, where $\hat{p}$ represents the causal program and $\hat{r}$ represents the corresponding natural language rationale. We prompt the LLM to internalize the algorithmic precision of the program $\hat{p}$ while articulating the reasoning steps in the natural language format of $\hat{r}$. Detailed inference prompts are in Appendix~\ref{app:prompt_reasoning}.

\section{Main Experiments of Python Program}
\label{sec:experimental_setup}

\subsection{Experimental Setup}
\paragraph{Benchmarks.}
We evaluate ECI2D on Math Reasoning: GSM8K~\cite{gsm8k}, SVAMP~\cite{svamp}, GSM-Symbolic~\cite{gsm-symbolic} and Logical Reasoning: DROP~\cite{drop}, Big-Bench Hard (BBH) logical subset (Date Understanding, Object Counting, and Word Sorting)~\cite{bbh}. Details of benchmarks are in Appendix~\ref{app:math_benchmark} and~\ref{app:logical_benchmark}. We report more results on other logical subsets of BBH in Appendix~\ref{app:more_datasets_in_logical_reasoning}.

\paragraph{Baselines and Implementation Details.}
We compare ECI2D against five baselines: Direct, Zero-shot CoT, Manual Few-shot CoT, Self-Generated Reinforced ICL, and Self-Generated Reinforced ICL with program-augmented demonstrations. We mainly employ two instruction models, \texttt{Meta-Llama-3-8B-Instruct}\footnote{https://huggingface.co/meta-llama/Meta-Llama-3-8B} and \texttt{Qwen2.5-7B-Instruct}\footnote{https://huggingface.co/Qwen/Qwen2.5-7B-Instruct}. Detailed settings for baselines and implementation can be found in Appendix~\ref{app:implementation_details}. We report more results of reasoning models in Appendix~\ref{app:more_experiment_math_results}, and compare against PoT~\cite{pot} in Appendix~\ref{app:pot}.

\begin{table}[t]
\centering
\caption{
  Ablation study analyzing different rationale components. We compare performance across three configurations: Program-only $\hat{p}$ (Prog.), Language rationale $\hat{r}$ alone (Lang.), and Both ($\hat{p}$ and $\hat{r}$). Best per model and benchmark in \textbf{bold}. GSM-S = GSM-Symbolic, P1/P2 = Plus-1/Plus-2.
}
\label{tab:ablation}

\setlength{\tabcolsep}{4pt}

\resizebox{\linewidth}{!}{
\begin{tabular}{l|ccc|ccc}
    \toprule
    \multirow{2}{*}{Benchmark} & \multicolumn{3}{c|}{\texttt{Llama3-8B}} & \multicolumn{3}{c}{\texttt{Qwen2.5-7B}} \\
    \cmidrule(lr){2-4}\cmidrule(lr){5-7}
     & Prog. & Lang. & Both & Prog. & Lang. & Both \\
    \midrule
    GSM8K    & 74.45 & 80.89 & \textbf{82.11} & 85.14 &  91.58  & \textbf{91.74} \\ 
    SVAMP    & 85.70 & 86.10 & \textbf{86.60} & 92.20 & 93.60 & \textbf{95.50} \\
    GSM-S    & 74.98 & 76.04 & \textbf{77.36} & 85.18 & 89.64 & \textbf{89.80} \\ 
    GSM-S P1 & 62.22 & 64.56 & \textbf{66.02} & 76.92 & 81.64 & \textbf{82.02} \\
    GSM-S P2 & 36.96 & 37.12 & \textbf{37.72} & 60.20 & 64.60 & \textbf{65.28} \\
    \midrule
    Mean     & 66.86 & 68.94 & \textbf{69.96} & 79.93 & 84.21 & \textbf{84.87} \\
    \bottomrule
\end{tabular}
}
\end{table}

\subsection{Main Results}
\label{sec:results}

\paragraph{Performance Comparison.} Table~\ref{tab:1} shows that while self-generated reinforced ICL
marginally improves upon manual prompts in some math datasets, it suffers from
degradation in logical reasoning scenarios. By contrast, ECI2D improves over it on
every benchmark, and its advantages are most pronounced in OOD scenarios, boosting
accuracy by over 7 points on the hardest GSM-Symbolic-P2 with Qwen2.5-7B where
other methods falter. Although manual Few-shot CoT retains the lead on parts of
BBH, owing to the domain gap where DROP centers on historical events unlike the
explicit date calculation in Date Understanding, ECI2D still consistently
outperforms Self-Generated Reinforced ICL there. 
% These results indicate that our
% interventional filtering distills faithful rationales that remain invariant across
% distribution shifts rather than improving in-distribution accuracy, where the
% margin falls within the seed-to-seed variation of distillation
% (Appendix~\ref{app:seeds}).
These results indicate that our interventional filtering distills faithful rationales invariant across distribution shifts: the GSM8K margin falls within the seed-to-seed variation of distillation (Appendix~\ref{app:seeds}), whereas the DROP and largest OOD gains far exceed it.

ECI2D also maintains high data efficiency, reducing the final demonstration counts
by 39\% to 57\% compared to the outcome-oriented baseline. This reduction implies
ECI2D effectively discards a large number of observationally correct but causally
spurious post-hoc examples. Subsampling that baseline to the same pool size
degrades it on every benchmark tested (Appendix~\ref{app:matched}), so the
compactness of $D_{\text{causal}}$ is a cost rather than a source of the
advantage. Transfer invariance is therefore a more reliable standard for
demonstration distillation than simple observational correctness.

\paragraph{Ablation Study.}
The ablation results in Table~\ref{tab:ablation} provide empirical evidence that
combining executable programs $\hat{p}$ with natural language rationales $\hat{r}$
yields a consistent performance benefit. The hybrid configuration utilizing both
$\hat{p}$ and $\hat{r}$ outperforms single-component setups on all ten
model--benchmark pairs, and the ordering is identical throughout.
The gap highlights the complementary roles of the two components. The Program-only
setting performs worse than the Language-only setting, reflecting a semantic
mismatch between code and natural language reasoning. Since the model answers in
natural language without executing the program, a demonstration containing only
$\hat{p}$ carries the computation but not the form in which the answer must be
produced. Nevertheless, the program remains necessary, as it supplies a
deterministic algorithmic scaffold that constrains the reasoning search to
invariant trajectories, while the language rationale provides the linguistic
mapping that translates those trajectories into expressive and interpretable
explanations. The \textit{+ Code} row of Table~\ref{tab:1} separates this format
from the filtering criterion, using the identical prompt as ECI2D but retaining
candidates by observational correctness alone. ECI2D outperforms it on all nine
benchmarks under both models, and adding code to the observational pipeline does not improve it consistently, so the gains stem from interventional verification rather
than from the presence of code.

\begin{table}[t]
\centering
\caption{
  Impact of sample number $N$ on performance. We report accuracy (\%) across benchmarks. The column `\textcolor{graytext}{Num}' denotes the size of the causal pool $|D_{\text{causal}}|$.
}
\label{tab:3}

\setlength{\tabcolsep}{4pt} % 微调列间距以适应 resizebox

\resizebox{\linewidth}{!}{
\begin{tabular}{l|cc|ccccc}
    \toprule
    % 表头行
    \multirow{2}{*}{Model} & \multirow{2}{*}{$N$} & \multirow{2}{*}{\textcolor{graytext}{Num}} & \multirow{2}{*}{GSM8K} & \multirow{2}{*}{SVAMP} & \multicolumn{3}{c}{GSM-Symbolic} \\
    \cmidrule(lr){6-8}
     & & & & & Base & P1 & P2 \\
    \midrule

    % --- Llama Block ---
    \multirow{3}{*}{\texttt{Llama}}  % 使用 texttt 保持字体一致
     & 3 & \textcolor{graytext}{3913} & \textbf{82.11} & 86.60 & 77.36 & \textbf{66.02} & 37.72 \\
     & 5 & \textcolor{graytext}{2248} & 79.61 & \textbf{87.20} & \textbf{78.38} & 64.28 & \textbf{37.80} \\
     & 8 & \textcolor{graytext}{1387} & 79.30 & 87.10 & 77.86 & 64.76 & 37.40 \\
    
    \midrule

    % --- Qwen Block ---
    \multirow{3}{*}{\texttt{Qwen}}  % 使用 texttt 保持字体一致
     & 3 & \textcolor{graytext}{4239} & \textbf{91.74} & \textbf{95.50} & \textbf{89.80} & \textbf{82.02} & \textbf{65.28} \\
     & 5 & \textcolor{graytext}{2417} & 87.79 & 89.70 & 83.46 & 73.74 & 56.32 \\
     & 8 & \textcolor{graytext}{1482} & 87.26 & 89.20 & 81.82 & 73.08 & 55.96 \\
     
    \bottomrule
\end{tabular}
}

\end{table}

\paragraph{Hyperparameter Sensitivity.}
We investigate how the count $N$ of logically similar samples affects the
distilled demonstrations. As shown in Table~\ref{tab:3}, performance does not
increase monotonically with $N$; the degradation is mild for \texttt{Llama3} but
reaches $8.96$ points on GSM-Symbolic-P2 for \texttt{Qwen2.5} when $N$ rises from
$3$ to $5$. This trend highlights a critical trade-off between verification rigor
and demonstration diversity. As $N$ increases, the retrieved samples inevitably include those of greater semantic distance, where minor structural deviations
cause valid causal skeletons to fail the strict execution check.

\paragraph{Robustness.}
\label{sec:error_propagation}
ECI2D relies on LLMs at three stages upstream of the filter, raising concerns about cascading error propagation. Retrieval may return a neighbor that does not share the reasoning structure,
abstraction may formalize the rationale incorrectly, and adaptation may alter the
skeleton or misbind the parameters of the target instance.
Appendix~\ref{app:human} quantifies the first two by human annotation, reporting high
agreement, and illustrates each adaptation failure. To
probe the third we deliberately degrade the Abstractor and Adapter on GSM8K
(\texttt{Qwen2.5-7B}), relaxing the prompt constraints and substituting
\texttt{Qwen2.5-1.5B}. As shown in Table~\ref{tab:error_propagation}, retention
falls from 57.3\% to 24.6\% while downstream accuracy remains above 87 points.
Upstream errors therefore cost coverage rather than pool quality, since any failed
check yields $S_\text{inv} = 0$ and discards the candidate.

\begin{table}[t]
\centering
\caption{Impact of Abstractor(Abs.)/Adapter(Adpt.) degradation on error propagation and downstream performance (GSM8K, Qwen2.5-7B). Ret. = Retention}
\label{tab:error_propagation}

\small % 保持原来的字体大小
\renewcommand{\arraystretch}{1.15}

% 关键1：把列间距压缩到 2pt
\setlength{\tabcolsep}{2pt} 

% 关键2：第一列使用 p{2.2cm}，强制长文字换行
\begin{tabular}{@{}p{2.2cm}cccc@{}}
\toprule
\textbf{Config} & \textbf{Abs. (\%)} & \textbf{Adpt. (\%)} & \textbf{Ret. (\%)} & \textbf{Acc (\%)} \\
\midrule
Standard   & 1.5  & 3.5  & 57.3 & 91.74 \\
W/o constraints& 6.3  & 12.2 & 41.3 & 88.95 \\
% 这里的文字会自动换行，不会溢出到右边
Qwen2.5-1.5B   & 10.6 & 25.1 & 24.6 & 87.48 \\
\bottomrule
\end{tabular}
\end{table}

\section{Additional Experiments of Hybrid Program}
\label{sec:experimental_additional}
\subsection{Experimental Setup}
\paragraph{Benchmarks.}
For the hybrid program abstraction of Section~\ref{sec:instantiation}, we use
CommonsenseQA~\cite{csqa} and BBH semantic subset (Sports Understanding and Ruin Names)~\cite{bbh}. Details of benchmarks are in Appendix~\ref{app:hybird_benchmark}.

\paragraph{Baselines and Implementation Details.}
We evaluate ECI2D with hybrid program with \texttt{Qwen2.5-7B-Instruct} against the
same baselines in Section~\ref{sec:experimental_setup}.
By default, we distill from the CommonsenseQA training set with $N=3$. More detailed settings for baselines and implementation can be found in Appendix~\ref{app:implementation_details}.

\subsection{Results}
Table~\ref{tab:instantiation} reports the hybrid abstraction on three benchmarks
that admit no deterministic program. Requiring one is harmful here, as the
\textit{+ Code} baseline falls below Self-Generated Reinforced ICL on two of the
three and below direct answering on CommonsenseQA. The hybrid abstraction improves
over both on all three and achieves the best accuracy on CommonsenseQA, exceeding
Self-Generated Reinforced ICL by $2.30$ points there and by $4.40$ and $3.60$
points on the two BBH subsets, while retaining 58.8\% of the training set against
79.1\% for the observational baseline. Manual Few-shot CoT retains the lead on BBH,
as in the main experiments, owing to the domain gap where the pool is distilled on
CommonsenseQA alone and transferred across domains. On Ruin Names this transfer
also falls below zero-shot inference. Of the interventions that score zero, 95\%
return a wrong option and the remainder fail to execute, so the criterion registers
whether the reasoning transfers rather than whether the program runs.
Appendix~\ref{app:instantiation} reports the full decomposition with the
constraints enforced on the abstraction and case study. The criterion
therefore extends beyond the domains where a deterministic program can be written.

\begin{table}[t]
\centering
\caption{Performance of the hybrid abstraction. The pool is distilled on
CommonsenseQA alone, so the two BBH subsets measure cross-domain transfer. Following standard CoT~\cite{cot}, we use
7-shot demonstrations on CommonsenseQA and 3-shot on BBH, and report accuracy
(\%). \textit{+ Code} restricts the abstraction to a deterministic Python program
and retains candidates by observational correctness alone. Best per benchmark in
\textbf{bold}, second-best \underline{underlined}. Sports = Sports Understanding,
Ruin = Ruin Names, Ret. = Retention.}
\label{tab:instantiation}
\small
\renewcommand{\arraystretch}{1.15}
\setlength{\tabcolsep}{4pt}
\begin{tabular}{@{}lcccc@{}}
\toprule
Method & CSQA & Sports & Ruin & Ret. \\
\midrule
Direct                   & 79.93 & 72.40 & 50.80 & -- \\
Zero-shot CoT            & 79.03 & 74.40 & \underline{53.60} & -- \\
Manual Few-shot CoT      & 79.52 & \textbf{86.80} & \textbf{64.00} & -- \\
Self-Gen. Reinforced ICL & \underline{80.34} & 74.80 & 43.60 & 79\% \\
\quad \textit{+ Code}             & 78.73 & 72.80 & 46.40 & 77\% \\
\rowcolor{blue!8} ECI2D-Hybrid (ours) & \textbf{82.64} & \underline{79.20} & 47.20 & 59\% \\
\bottomrule
\end{tabular}
\end{table}

\section{Conclusion}
In this paper, we addressed the critical challenge of \textit{observational equivalence} in self-generated In-Context Learning (ICL), where standard observation-based filtering fails to distinguish between faithful causal rationales and spurious post-hoc rationales. We argued that a faithful relation should remain effective across similar environments, implying \textit{transfer invariance}. To instantiate this, we introduced the Executable Causal Intervention for Demonstrations Distillation (ECI2D) framework. By abstracting natural language rationales into executable programs and applying the $do$-operator via causal interventions, ECI2D shifts the demonstration selection paradigm from passive observation to active intervention. ECI2D outperforms outcome-oriented reinforced ICL by over 7 points while retaining fewer demonstrations, with gains scaling under increasing distribution shift. These results establish intervention-based invariance as a sound criterion for distilling faithful rationales in self-generated In-Context Learning.

\section*{Limitations}

\textbf{Computational cost.} ECI2D adds overhead during offline distillation: iterative rationale sampling, program abstraction and adaptation, neighbor  retrieval, and repeated sandbox execution. Constructing $D_{\text{causal}}$ therefore costs more than outcome-oriented filtering, although the cost is amortized offline and inference is unaffected (Appendix~\ref{app:cost}).

\textbf{Availability of logically similar samples.} Verification requires $N$ instances sharing a candidate's logical skeleton, which ECI2D obtains by retrieval from the training set. The criterion is thus applicable only where such instances exist, and degrades when the data are scarce or problem structures diverse. Section~\ref{sec:results} exhibits the dependency: the pool shrinks from $3{,}913$ to $1{,}387$ candidates on GSM8K as $N$ rises from $3$ to $8$, so genuinely same-structure instances are not abundant even at this scale. Constructing the neighborhood from the source instance rather than retrieving it would remove the dependency, which we leave to future work.

\textbf{Incomplete identification.} Transfer invariance is a necessary but not sufficient condition on faithfulness (Appendix~\ref{app:icm_proof}). It rules out latent factors that vary across $\mathcal{N}(x_i)$, such as memorized $(x,y)$ associations and surface-level shortcuts, but not factors that act uniformly within the verified neighborhood. ECI2D should therefore be read as a conservative local filter rather than an identifier of the causal mechanism. Its residual error is asymmetric in direction. Abstraction and adaptation failures cost coverage (Section~\ref{sec:error_propagation}), whereas silent repair by the abstractor can admit an unfaithful rationale and is not measured here (Appendix~\ref{app:human}).

% Bibliography entries for the entire Anthology, followed by custom entries
%\bibliography{custom,anthology-overleaf-1,anthology-overleaf-2}

% Custom bibliography entries only
\bibliography{custom}

\appendix

\begin{table*}[t]
\centering
\caption{More results on math reasoning benchmarks using \texttt{DeepSeek-R1-Distill-Qwen-7B}. We utilize 8-shot demonstrations.
We reproduce all baselines and report accuracy across all benchmarks. The demonstration retention rate after distillation is also reported.
The best results for each setting are marked in \textbf{bold}.}
\label{tab:deepseek_comparison}
\small
\renewcommand{\arraystretch}{1.15}
\setlength{\tabcolsep}{3.5pt}
\begin{tabular}{l c c c c c c}
\toprule
Method & GSM8K & SVAMP & GSM-Symbolic & GSM-Symbolic-P1 & GSM-Symbolic-P2 & Retention \\
\midrule
Direct & 42.30 & 58.80 & 72.74 & 69.76 & 60.20 & -- \\
Zero-shot CoT & 80.26 & 81.60 & 74.78 & 72.84 & 66.08 & -- \\
Manual Few-shot CoT  & 82.26 & 84.80 & 82.70 & 80.08 & 72.72 & -- \\
Self-Gen. Reinforced ICL  & 88.44 & 88.40 & 89.30 & 84.94 & 78.24 & 91\% \\
\rowcolor{blue!8} ECI2D (Ours) & \textbf{92.04} & \textbf{92.40} & \textbf{91.94} & \textbf{89.32} & \textbf{79.72} & 53\% \\
\bottomrule
\end{tabular}
\end{table*}

\begin{table*}[t]
\centering
\caption{More results on math reasoning benchmarks using \texttt{Qwen3-8B}. Following~\cite{qwen3technicalreport}, we utilize 4-shot demonstrations.
We reproduce all baselines and report accuracy across all benchmarks. The demonstration retention rate after distillation is also reported.
The best results for each setting are marked in \textbf{bold}.}
\label{tab:qwen3_main}
\small
\renewcommand{\arraystretch}{1.15}
\setlength{\tabcolsep}{3.5pt}
\begin{tabular}{l c c c c c c}
\toprule
Method & GSM8K & SVAMP & GSM-Symbolic & GSM-Symbolic-P1 & GSM-Symbolic-P2 & Retention Rate \\
\midrule
Direct & 73.09 & 62.60 & 74.28 & 71.96 & 64.80 & -- \\
Zero-shot CoT & 87.88 & 80.00 & 85.62 & 76.88 & 70.16 & -- \\
Manual Few-shot CoT  & 84.42 & 74.60 & 79.14 & 82.76 & 72.48 & -- \\
Self-Gen. Reinforced ICL  & 89.95 & 89.60 & 90.36 & 83.40 & 80.52 & 88\% \\
\rowcolor{blue!8} ECI2D (Ours) & \textbf{94.54} & \textbf{93.20} & \textbf{93.80} & \textbf{88.12} & \textbf{82.24} & 54\% \\
\bottomrule
\end{tabular}
\end{table*}

\section{More Experiment Results}
\label{app:more_experiment_results}
\subsection{More models in Math Reasoning}
\label{app:more_experiment_math_results}
While our primary experiments align with standard self-generated ICL protocols using instruction-tuned LLMs, we extend our evaluation to reasoning-specialized architectures to assess the architectural generalizability of ECI2D. Specifically, we conduct math reasoning benchmarks on \texttt{DeepSeek-R1-Distill-Qwen-7B}\footnote{https://huggingface.co/deepseek-ai/DeepSeek-R1-Distill-Qwen-7B} and \texttt{Qwen3-8B}\footnote{https://huggingface.co/Qwen/Qwen3-8B}. As detailed in Table~\ref{tab:deepseek_comparison} and Table~\ref{tab:qwen3_main}, ECI2D surpasses the self-generated reinforced ICL baseline on all evaluated benchmarks. Notably, this advantage is achieved despite a lower demonstration retention rate, suggesting that the distilled subset prioritizes fidelity over volume. Two observations follow. First, the intervention-based criterion transfers across the four models we evaluate, including architectures explicitly optimized for chain-of-thought generation. Second, the persistent gap suggests that reasoning-specialized models also produce rationales which satisfy outcome-oriented verification without transferring to logically similar samples.

\subsection{More datasets in Logical Reasoning}
\label{app:more_datasets_in_logical_reasoning}
We extend our analysis to the Boolean Expressions and Penguins in a Table subsets. Semantic subsets of BBH are covered by the hybrid program of Section~\ref{sec:experimental_additional} and Appendix~\ref{app:instantiation}. As detailed in Table~\ref{tab:add_logical}, ECI2D achieves performance comparable to manual prompts while surpassing the self-generated reinforced ICL baseline on both subsets and under both models, which indicates that transfer-invariant distillation extends to logical benchmarks beyond those used in our main experiments.

\begin{table*}[t]
\centering
\caption{More results on logical reasoning benchmarks across different models. The best results for each setting are marked in \textbf{bold}, and the second-best results are \underline{underlined}.}
\label{tab:add_logical}
\small
\renewcommand{\arraystretch}{1.15}
\setlength{\tabcolsep}{3.5pt}
\begin{tabular}{l l c c}
\toprule
Model & Method & Boolean Expressions & Penguins in a Table \\
\midrule
\multirow{3}{*}{\texttt{Llama3}} & Manual Few-shot CoT & \underline{72.40} & \underline{47.15} \\
 & Self-Gen. Reinforced ICL & 68.40 & 44.52 \\
\rowcolor{blue!8} & ECI2D (Ours) & \textbf{76.00} & \textbf{47.26} \\
\midrule
\multirow{3}{*}{\texttt{Qwen2.5}} & Manual Few-shot CoT & 83.60 & \textbf{47.58} \\
 & Self-Gen. Reinforced ICL & \underline{87.20} & 46.58 \\
\rowcolor{blue!8} & ECI2D (Ours) & \textbf{91.60} & \underline{47.13} \\
\bottomrule
\end{tabular}
\end{table*}

\subsection{Stability across Distillation Seeds}
\label{app:seeds}
Inference uses greedy decoding ($\tau=0.0$) and is therefore deterministic, but constructing $D_{\text{causal}}$ relies on sampling ($\tau=0.7$) and is seed-dependent. We rebuild the causal pool from scratch under three independent seeds ($42$, $43$, $44$) and report the resulting distribution in Table~\ref{tab:seeds}; Table~\ref{tab:1} reports the run with seed $42$. Variation is below $0.7$ points on six of the nine benchmarks and below $2.5$ points on all of them, and retention rates vary by less than $0.4$ points, indicating that the composition of the distilled pool is stable rather than an artifact of a favorable sample. The comparison also delimits which margins in Table~\ref{tab:1} are meaningful: the GSM8K margins (+0.23 and +0.16) are of the same order as this variation, whereas the gains on DROP (+5.88 and +7.81) and under distribution shift (e.g., +7.08 and +7.16 on GSM-Symbolic-P1/P2 with Qwen2.5) exceed it by one to two orders of magnitude.

\begin{table}[t]
\centering
\caption{
  Stability of ECI2D across three independent distillation seeds ($42$, $43$, $44$), with the causal pool $D_{\text{causal}}$ rebuilt from scratch for each seed. \textcolor{graytext}{Seed 42} reproduces the run reported in Table~\ref{tab:1}. Ret. = Retention.
}
\label{tab:seeds}

\setlength{\tabcolsep}{4pt}

\resizebox{\linewidth}{!}{
\begin{tabular}{l|cc|cc}
    \toprule
    \multirow{2}{*}{Benchmark} & \multicolumn{2}{c|}{\texttt{Llama3-8B}} & \multicolumn{2}{c}{\texttt{Qwen2.5-7B}} \\
    \cmidrule(lr){2-3}\cmidrule(lr){4-5}
     & \textcolor{graytext}{Seed 42} & 3 seeds & \textcolor{graytext}{Seed 42} & 3 seeds \\
    \midrule
    GSM8K    & \textcolor{graytext}{82.11} & 82.36\,\scriptsize{$\pm$0.23} & \textcolor{graytext}{91.74} & 91.96\,\scriptsize{$\pm$0.23} \\
    SVAMP    & \textcolor{graytext}{86.60} & 86.70\,\scriptsize{$\pm$0.36} & \textcolor{graytext}{95.50} & 95.53\,\scriptsize{$\pm$0.65} \\
    GSM-S    & \textcolor{graytext}{77.36} & 77.35\,\scriptsize{$\pm$0.13} & \textcolor{graytext}{89.80} & 89.79\,\scriptsize{$\pm$0.12} \\
    GSM-S P1 & \textcolor{graytext}{66.02} & 66.05\,\scriptsize{$\pm$0.04} & \textcolor{graytext}{82.02} & 82.08\,\scriptsize{$\pm$0.18} \\
    GSM-S P2 & \textcolor{graytext}{37.72} & 37.88\,\scriptsize{$\pm$0.16} & \textcolor{graytext}{65.28} & 65.40\,\scriptsize{$\pm$0.14} \\
    \midrule
    DROP     & \textcolor{graytext}{80.87} & 81.00\,\scriptsize{$\pm$0.62} & \textcolor{graytext}{80.54} & 80.89\,\scriptsize{$\pm$0.37} \\
    Date U.  & \textcolor{graytext}{62.80} & 63.47\,\scriptsize{$\pm$1.51} & \textcolor{graytext}{78.00} & 77.87\,\scriptsize{$\pm$1.41} \\
    Obj. C.  & \textcolor{graytext}{80.00} & 79.87\,\scriptsize{$\pm$1.40} & \textcolor{graytext}{85.20} & 85.47\,\scriptsize{$\pm$1.22} \\
    Word S.  & \textcolor{graytext}{59.60} & 60.53\,\scriptsize{$\pm$1.01} & \textcolor{graytext}{24.40} & 26.53\,\scriptsize{$\pm$2.45} \\
    \midrule
    Ret. (math)  & \textcolor{graytext}{52.00} & 51.97\,\scriptsize{$\pm$0.39} & \textcolor{graytext}{57.00} & 57.21\,\scriptsize{$\pm$0.27} \\
    Ret. (logic) & \textcolor{graytext}{27.00} & 27.25\,\scriptsize{$\pm$0.27} & \textcolor{graytext}{32.00} & 32.16\,\scriptsize{$\pm$0.17} \\
    \bottomrule
\end{tabular}
}

\end{table}

\subsection{Size-Matched Observational Pool}
\label{app:matched}
The causal pool is smaller than the observational one, which raises the question of whether the improvement follows from pool size rather than from pool composition. To separate the two, we uniformly subsample the observational pool to $|D_{\text{causal}}|$ and repeat inference with identical retrieval, prompt format, shot count and decoding. The subsample is drawn from the observationally filtered pool and is not constrained to lie within $D_{\text{causal}}$.

As shown in Table~\ref{tab:matched}, reducing the pool degrades the observational baseline on all three benchmarks, by $2.73$ points on GSM8K, $1.68$ on GSM-Symbolic-P2 and $1.45$ on DROP. Pool size therefore acts in the opposite direction to the one the concern presupposes. The compactness of $D_{\text{causal}}$ is a cost that ECI2D absorbs rather than a source of its advantage, and at equal pool size the margin over the observational criterion widens on every benchmark.

Random subsampling is the appropriate null model here because it preserves the composition of the observational pool in expectation, so any difference from the full pool is attributable to size alone. It is not, however, the strongest selection heuristic available at that size, and the comparison should be read as isolating the effect of pool size rather than as a comparison against demonstration selection methods in general.

\begin{table}[t]
\centering
\caption{Size-matched control with \texttt{Meta-Llama-3-8B-Instruct}. The observational pool is subsampled from $6{,}427$ to $3{,}913$ candidates for math reasoning and from $2{,}981$ to $1{,}278$ for DROP, matching $|D_{\text{causal}}|$ in each domain. Retrieval, shot count, prompt format and decoding are identical throughout.}
\label{tab:matched}
\small
\renewcommand{\arraystretch}{1.15}
\setlength{\tabcolsep}{4pt}
\begin{tabular}{@{}l ccc@{}}
\toprule
Method & GSM8K & GSM-S P2 & DROP \\
\midrule
Self-Gen. Reinforced ICL & 81.88 & 36.92 & 74.99 \\
\quad @ matched pool size & 79.15 & 35.24 & 73.54 \\
\rowcolor{blue!8} ECI2D (ours) & \textbf{82.11} & \textbf{37.72} & \textbf{80.87} \\
\bottomrule
\end{tabular}
\end{table}

\subsection{Comparison with Program-Aided Prompting}
\label{app:pot}
PAL and PoT differ primarily in prompt format rather than in mechanism, and we take PoT as representative of program-aided prompting. PoT~\cite{pot} generates a program for each test query and executes it to obtain the answer, whereas ECI2D uses programs only during offline distillation and issues a standard few-shot prompt at inference time. The two operate under different deployment assumptions, and we report the comparison separately for that reason rather than folding it into Table~\ref{tab:1}. We restrict it to math reasoning: on the BBH subsets the task reduces to a single library call once a program is emitted, so a comparison there measures interpreter access rather than demonstration quality.

As shown in Table~\ref{tab:pot}, access to program execution improves over human-written prompts under the training distribution ($81.65$ against $79.00$ on GSM8K) but does not confer robustness to structural complexity: PoT reaches only $58.62$ on GSM-Symbolic-P1 and $32.28$ on P2, below the self-generated baseline and $7.40$ and $5.44$ points below ECI2D. The bottleneck under distribution shift therefore lies in selecting the reasoning structure rather than in executing it precisely, and interpreter access does not substitute for filtering which rationales enter the demonstration pool.

\begin{table}[t]
\centering
\caption{Comparison with program-aided prompting on math reasoning (\texttt{Meta-Llama-3-8B-Instruct}). PoT$^\dagger$ generates and executes a program for each test query and therefore requires interpreter access at inference time; the remaining methods issue a standard few-shot prompt. No ranking is marked, since the two regimes operate under different deployment assumptions. GSM-S = GSM-Symbolic, P1/P2 = Plus-1/Plus-2.}
\label{tab:pot}
\setlength{\tabcolsep}{4pt}
\resizebox{\linewidth}{!}{
\begin{tabular}{l ccccc}
\toprule
Method & GSM8K & SVAMP & GSM-S & P1 & P2 \\
\midrule
Manual Few-shot CoT & 79.00 & 81.30 & 75.32 & 58.54 & 32.04 \\
Self-Gen. Reinforced ICL & 81.88 & 84.80 & 76.88 & 63.74 & 36.92 \\
PoT$^\dagger$ & 81.65 & 85.20 & 75.44 & 58.62 & 32.28 \\
\rowcolor{blue!8} ECI2D (ours) & 82.11 & 86.60 & 77.36 & 66.02 & 37.72 \\
\bottomrule
\end{tabular}
}
\end{table}

\section{Benchmark Details}
\label{app:datasets}

In this section, we provide detailed specifications of the datasets used in our
experiments. We categorize them into Math Reasoning, Logical Reasoning, and Semantic
Reasoning benchmarks. Specifically, GSM8K, a filtered subset of DROP, and CommonsenseQA serve
as In-Distribution (ID) datasets for distilling and evaluation, while SVAMP,
GSM-Symbolic, and Big-Bench Hard (BBH) serve as Out-Of-Distribution (OOD) benchmarks
to evaluate the robustness and effectiveness of our proposed framework. All the
distilled demonstrations are from training set, no test examples are included in
$\mathcal{N}(x)$ or $D_\text{causal}$.

\subsection{Math Reasoning Benchmarks}
\label{app:math_benchmark}

\textbf{GSM8K (In-Distribution)}~\cite{gsm8k} consists of 8.5K high-quality linguistically diverse grade school math word problems. It is divided into 7,473 training examples and 1,319 test examples. Each problem takes between 2 to 8 reasoning steps to solve. We use the training set of GSM8K to construct our invariant causal demonstrations and the test set for standard performance evaluation.

\textbf{SVAMP (Out-Of-Distribution)}~\cite{svamp} is a challenge dataset created by applying varying structural modifications to elementary math word problems. It consists of 1,000 examples designed to test the robustness of models against changes in word order and problem structure. In our experiments, SVAMP is used as an OOD benchmark to verify whether the distilled rationales can generalize to problems with different linguistic patterns than GSM8K.

\textbf{GSM-Symbolic (Out-Of-Distribution)}~\cite{gsm-symbolic} is an improved benchmark created from symbolic templates that allow for the generation of a diverse set of questions. Unlike static datasets, it allows for the generation of massive variations of a single problem logic. We utilize these three different difficulty level subsets to evaluate robustness: \textit{GSM-Symbolic} (Standard), \textit{GSM-Symbolic-Plus-1}, and \textit{GSM-Symbolic-Plus-2}. 
GSM-Symbolic contains 100 templates generating 5,000 examples. It mirrors the distribution of the original GSM8K but with refreshed numbers and names to prevent memorization. One or two clauses can be added to the questions to increase their difficulty, yielding GSM-Symbolic-Plus-1 (GSM-Symbolic-P1) and GSM-Symbolic-Plus-2 (GSM-Symbolic-P2), respectively.

\subsection{Logical Reasoning Benchmarks}
\label{app:logical_benchmark}
\textbf{Discrete Reasoning Over Paragraphs (DROP) (In-Distribution)}~\cite{drop} is a reading comprehension benchmark requiring discrete reasoning over paragraphs. The original dataset contains 77,409 training and 9,536 validation examples. DROP has nine reasoning categories, covering discrete reasoning and span information extraction. To save computational resources and improve distilling efficiency, we pre-filter the benchmark to retain only questions requiring \textit{sorting}, \textit{counting}, and \textit{temporal} reasoning, resulting 4,732 training and 737 validation examples. We developed a heuristic classifier based on question keywords and answer types. The filtering process is formally described in Algorithm~\ref{alg:drop_filter}.

\textbf{Big-Bench Hard (BBH) (Out-Of-Distribution)}~\cite{bbh} is a suite of 23 challenging tasks from BIG-Bench where language models previously struggled. To evaluate logical reasoning, we select three specific subsets that require multi-step logical deduction. \textit{Date Understanding}: involves inferring dates from relative time descriptions (e.g., "yesterday", "the first Monday of April"). \textit{Object Counting}: requires maintaining a mental state to count objects satisfying specific conditions. \textit{Word Sorting}: tests the ability to lexicographically sort a list of words. \textit{Boolean Expressions}: evaluates formal logical reasoning by requiring models to determine the truth value of nested Boolean expressions containing operators such as \textsc{and}, \textsc{or}, and \textsc{not}. \textit{Penguins in a Table}: assesses structured data comprehension by querying a tabular dataset of penguin attributes, requiring models to perform precise filtering, counting, and comparison operations to derive deterministic answers. Based on this classification, we extracted a balanced subset focusing on quantitative and temporal logic to serve as the training source for our logical reasoning experiments.

\subsection{Semantic Reasoning Benchmarks}
\label{app:hybird_benchmark}
\textbf{CommonsenseQA (In-Distribution)}~\cite{csqa} is a five-way multiple-choice
benchmark built from ConceptNet, in which each question targets a commonsense
relation between concepts. It contains 9,741 training and 1,221 validation examples,
and the test labels are withheld for the official leaderboard, so we evaluate on the
validation split. Answering a question requires world knowledge about objects and
actions rather than computation, and no deterministic program decides it. We distil
from the full training split and use CommonsenseQA as the ID benchmark for the hybrid
abstraction of Section~\ref{sec:instantiation}.

\textbf{BBH Semantic Subset (Out-Of-Distribution)}~\cite{bbh} comprises the two
subsets of BBH whose answers likewise rest on world knowledge, each containing 250
examples. \textit{Sports Understanding}: judges whether an artificially constructed
sentence about sports is plausible, which requires knowing which sport an athlete
plays and which actions belong to that sport. \textit{Ruin Names}: selects the
humorous edit of an artist or movie name among near-identical alternatives, which
requires recognising wordplay. Neither is expressible as a deterministic program.
Both are evaluated with the pool distilled on CommonsenseQA, so they measure
cross-domain transfer in the same way that BBH does for DROP.

\begin{algorithm}[!t] % !t 强制优先置顶，避免沉底
\caption{Invariant Reasoning Distillation of ECI2D}
\label{alg:ECI2D_distillation}
\begin{algorithmic}[1] % \small 缩小字号，[1] 保留行号
    \Require Training set $\mathcal{D} = \{(x_i, y_i)\}$, LLM Generator $\Theta_{\text{Gen}}$, LLM Abstractor $\Theta_{\Pi}$, LLM Adapter $\Theta_{\text{Adapter}}$, threshold $\delta$
    \Ensure Invariant Causal Demonstrations $D_{\text{causal}}$
    \State Initialize $D_{\text{causal}} \leftarrow \emptyset$
    \For{each sample $(x_i, y_i) \in \mathcal{D}$}
        \State \textbf{// Step 1: Causal Abstraction}
        \State Generate rationale $\hat{r}_i$ for $x_i$ using $\Theta_{\text{Gen}}$
        \State Abstract into executable program: $\hat{p}_i \leftarrow \Theta_\Pi(\hat{r}_i)$
        \State Initialize cumulative score $S_{\mathrm{sum}} \leftarrow 0$
        \For{each neighbor $(x_j, y_j) \in \mathcal{N}(x_i)$}
            \State \textbf{// Step 2: Executable Causal Intervention}
            \State Adapt: $\hat{p}_{i\to j} \leftarrow \Theta_{\text{Adapter}}(\hat{p}_i, x_j)$
            \State Execute: $\hat{y}_j \leftarrow \text{Exec}(\hat{p}_{i\to j})$
            \State \textbf{// Step 3: Transfer Invariance Verification}
            \State Score: $S_j \leftarrow S_\text{inv}(x_j, do(\hat{r}_i), \hat{y}_j)$
            \State $S_{\mathrm{sum}} \leftarrow S_{\mathrm{sum}} + S_j$
        \EndFor
        \State Compute average score: $S_{\mathrm{avg}} \leftarrow \frac{S_{\mathrm{sum}}}{|\mathcal{N}(x_i)|}$
        \If{$S_{\mathrm{avg}} \ge \delta$}
            \State Add $(x_i, \hat{p}_i, \hat{r}_i, \hat{y}_i)$ to $D_{\text{causal}}$
        \EndIf
    \EndFor
    \State \Return $D_{\text{causal}}$
\end{algorithmic}
\end{algorithm}

\begin{algorithm}[tb]
\caption{Heuristic Filtering for DROP Logical Subsets}
\label{alg:drop_filter}
\begin{algorithmic}
\Require Question string $Q$, Answer types set $\mathcal{A}$
\Ensure Logical Category $C \in \{\text{Sorting, Counting, Temporal, Other}\}$

\State \textbf{// 1. Identify Sorting Logic}
\If{any keyword in ["first", "last", "most", "least", "largest", "smallest", "order", "sort"] exists in $Q$}
    \State \textbf{return} $C \leftarrow \text{Sorting}$
\EndIf

\State \textbf{// 2. Identify Counting Logic}
\If{"how many" is in $Q$}
    \If{no arithmetic keyword (e.g., "difference", "change", "total", "sum", "more", "less") exists in $Q$}
        \If{"number" is in $\mathcal{A}$}
            \State \textbf{return} $C \leftarrow \text{Counting}$
        \EndIf
    \EndIf
\EndIf

\State \textbf{// 3. Identify Temporal Logic}
\If{"date" is in $\mathcal{A}$}
    \State \textbf{return} $C \leftarrow \text{Temporal}$
\EndIf
\If{any keyword in ["year", "month", "day", "date", "period", "when"] exists in $Q$}
    \State \textbf{return} $C \leftarrow \text{Temporal}$
\EndIf

\State \textbf{return} $C \leftarrow \text{Other}$ \Comment{Discarded in our subset}
\end{algorithmic}
\end{algorithm}

\section{Baselines and Implementation Details}
\label{app:implementation_details}

In this section, we provide a comprehensive overview of the baseline methods, model specifications, and the specific implementation configurations used in our experiments. Code and data-generation scripts will be released upon acceptance.

\subsection{Baseline Methods}
To evaluate the effectiveness and robustness of ECI2D, we compare it against a spectrum of prompting paradigms ranging from standard usage to state-of-the-art self-generated methods.

\textbf{Direct Prompting}: The model predicts the answer directly given the question, without any intermediate reasoning steps. This serves as the lower bound to assess the necessity of rationales.

\textbf{Zero-Shot CoT}~\cite{kojima2022large}: We append the trigger prompt \textit{Let's think step by step} to the input. This baseline evaluates the  intrinsic reasoning capability of LLMs without the guidance of external demonstrations.

\textbf{Manual Few-Shot CoT}~\cite{cot}: We utilize high-quality human-annotated Chain-of-Thought demonstrations to serve as a strong baseline. To ensure a fair comparison with established standards, we source the demonstrations from representative works, ~\cite{cot} for all math reasoning datasets (GSM8K, SVAMP, GSM-Symbolic), ~\cite{leasttomost} for the DROP dataset, and~\cite{bbh} for BBH tasks.

\textbf{Self-Generated Reinforced ICL}~\cite{from_few_to_many,many_shot,are_human_gen,self_gen_icl,auto_cot}: This is the primary baseline to our approach. It distills demonstrations based on an outcome-oriented criterion, retaining generated rationales that lead to the correct ground truth answer. Aside from the difference in the quality of the self-generated demonstrations, our method maintains consistency with self-generated reinforced ICL in terms of retrieval ICL, Sentence-transformer selection, and hyperparameter settings. Comparing against this baseline highlights the specific contribution of our transfer intervention mechanism over simple correctness distilling.

\textbf{Self-Generated Reinforced ICL + Code}: A control that separates the filtering criterion from the demonstration format. It uses the identical generation prompt as ECI2D (Appendix~\ref{app:prompt_gen_abs}), so each candidate contains both a natural-language rationale and a Python program, but retains candidates by observational correctness on the source question alone, omitting Steps 2 and 3 of Algorithm~\ref{alg:ECI2D_distillation}. Retrieval, encoder and all hyperparameters are identical to ECI2D. The same control is applied to the hybrid abstraction of
Section~\ref{sec:instantiation}, which uses the template of
Appendix~\ref{app:prompt_hybrid_abstraction} without the declaration block.

\textbf{PoT}~\cite{pot}: The model emits a Python program for the test query, which an external interpreter executes to obtain the answer. Unlike all other methods considered here, this requires interpreter access at test time; the comparison is reported in Appendix~\ref{app:pot}.

All the baselines adopt a greedy decoding configuration to ensure consistency and reproducibility. This hierarchy of baselines allows us to deconstruct the performance gains, in which improving over Zero-Shot proves the value of demonstrations; improving over Self-generated Reinforced ICL proves the value of intervention over observation.

% \subsection{Model Selection}
% We conduct experiments on two open-source Large Language Models.

% \textbf{Meta-Llama-3-8B-Instruct}\footnote{https://huggingface.co/meta-llama/Meta-Llama-3-8B}: As a widely adopted encoder-only model, Llama-3 exhibits strong instruction-following capabilities and serves as a representative benchmark for the 8B parameter scale.

% \textbf{Qwen2.5-7B-Instruct}\footnote{https://huggingface.co/Qwen/Qwen2.5-7B-Instruct}: We specifically select Qwen2.5 due to its specialized strengths in math reasoning and code generation. Since our ECI2D framework relies on abstracting reasoning into Python programs, Qwen2.5's coding proficiency makes it an ideal candidate to test the limits of our framework.

% All models are deployed using the vLLM~\cite{vllm} library on a single NVIDIA A6000 (48GB) GPU to ensure efficient, high-throughput inference.

\subsection{Model Selection}
We conduct experiments on four open-source Large Language Models, spanning standard instruction-tuned LLMs and reasoning-specialized architectures.

\textbf{Meta-Llama-3-8B-Instruct}\footnote{https://huggingface.co/meta-llama/Meta-Llama-3-8B-Instruct}: As a widely adopted decoder-only model, Llama-3 exhibits strong instruction-following capabilities and serves as a representative benchmark for the 8B parameter scale.

\textbf{Qwen2.5-7B-Instruct}\footnote{https://huggingface.co/Qwen/Qwen2.5-7B-Instruct}: We specifically select Qwen2.5 due to its specialized strengths in math reasoning and code generation. Since our ECI2D framework relies on abstracting reasoning into Python programs, Qwen2.5's coding proficiency makes it an ideal candidate to test the limits of our framework.

\textbf{DeepSeek-R1-Distill-Qwen-7B}\footnote{https://huggingface.co/deepseek-ai/DeepSeek-R1-Distill-Qwen-7B}: Distilled from DeepSeek-R1 onto a Qwen2.5-7B backbone, this model embeds long chain-of-thought reasoning patterns directly into its parameters. We include it to assess whether ECI2D's intervention-based filtering remains effective on architectures explicitly optimized for extended rationale generation, where the risk of post-hoc rationalization is arguably higher.

\textbf{Qwen3-8B}\footnote{https://huggingface.co/Qwen/Qwen3-8B}: The latest 8B model in the Qwen series, featuring further-strengthened reasoning and coding capabilities relative to Qwen2.5. We include it to verify that the gains of ECI2D persist as the underlying base model continues to improve, rather than reflecting weaknesses specific to a particular generation of LLMs.

All models are deployed using the vLLM~\cite{vllm} library on a single NVIDIA A6000 (48GB) GPU to ensure efficient, high-throughput inference.

\subsection{Implementation Details}

\paragraph{Logically Similar Sample Retrieval for Intervention}
To construct the logically similar samples $\mathcal{N}(x)$ for the executable causal intervention, we employ a coarse-to-fine retrieval strategy. This two-stage process is designed to ensure that the retrieved neighborhood $\mathcal{N}(x)$ satisfies both feasibility and comparability for the subsequent intervention. In the initial coarse retrieval stage, we utilize the \texttt{sentence-transformers/all-MiniLM-L6-v2}\footnote{https://huggingface.co/sentence-transformers/all-MiniLM-L6-v2} model to encode input queries. For each training sample $x$, we retrieve a broad candidate pool of size $5N$. This step serves as a necessary domain constraint, ensuring that the candidate samples reside within the same semantic context as the input. Subsequently, in the fine-grained selection stage, we further leverage the LLM to identify logically analogous examples. Since mere semantic similarity does not guarantee logical compatibility, and applying a rationale to a logically heterogeneous problem would render the verification meaningless, we prompt the LLM to analyze the underlying problem logic. It selects the final top-$N$ (with $N \in \{3, 5, 8\}$) samples that share similar reasoning logic with the input $x$. This ensures that the intervention is executable and the results are comparable, thereby providing a testbed for verifying the transfer invariance of the distilled rationales. The prompts for fine-grained retrieval can be found in Appendix~\ref{app:prompt_retriveal}.

\paragraph{Invariance Threshold Settings}
The transfer invariance verification relies on a scoring threshold $\delta$ to determine if a rationale's logic holds across logically similar samples.
For math reasoning, since math problems have unique, objective solutions, we set the threshold $\delta = 1.0$. A rationale is considered invariant only if the executed program yields the exact match of the ground truth for the neighbor problem. For logical reasoning (DROP), the DROP dataset often involves answer sets or text spans where slight formatting variations are permissible. We use the F1 score as the evaluation metric and set the invariance threshold $\delta = 0.9$. This strict yet flexible threshold ensures that the logic is correct while accommodating minor textual discrepancies common in reading comprehension tasks. For semantic reasoning (CommonsenseQA), the answer is a single option from a closed
label set, so we again set $\delta = 1.0$ and retain a candidate only when all $N$
interventions recover the ground truth of their target.

\paragraph{Hyperparameter Configuration}
We adopt a distinct temperature strategy for the distillation and inference phases to balance diversity and determinism. In invariant reasoning distillation phase, we set temperature $T=0.7$. A higher temperature is crucial here to encourage the model to explore a diverse space of rationales. This diversity is necessary to uncover potentially faithful causal mechanisms among various spurious possibilities. In causal steering inference phase, we set $T=0.0$ (greedy decoding). This ensures reproducibility and minimizes variance, forcing the model to strictly adhere to the guidance provided by the selected causal demonstrations without introducing stochastic noise.

\paragraph{Timeout in Execution}
For the causal abstraction, we limit the generation to a maximum of 512 tokens to encourage concise algorithmic logic. All generated programs are executed in a sandboxed Python environment with a timeout of 2 seconds to prevent infinite loops.

\paragraph{Hybrid Program Configuration}
The hybrid abstraction of Section~\ref{sec:instantiation} is evaluated with
\texttt{Qwen2.5-7B-Instruct}. We distil from the full CommonsenseQA training set of
9,741 examples and retrieve logically similar samples with
\texttt{all-mpnet-base-v2}, setting $N = 3$ and $\delta = 1.0$ as in the preceding
paragraph.
 The coarse retrieval pool is $5N$. $\Theta_\Pi$ decodes at temperature $0.7$ and every
other call is greedy, following the main experiments. $\Theta_\Pi$ is allowed two
retries when a candidate fails the constraints of
Appendix~\ref{app:instantiation}, while $\Theta_{\text{Adapter}}$ and
$\Theta_{\text{Eval}}$ are each called once. Every execution is bounded by a two
second timeout. At inference time we use 7-shot demonstrations on CommonsenseQA and
3-shot on the semantic subset of BBH.

CommonsenseQA presents five candidate answers, so $\Theta_\Pi$ abstracts a rationale
that discriminates among them into a test of the candidates against the properties
the rationale names, and $\mathrm{s}_i$ visits each candidate in turn. One
intervention therefore issues one call to $\Theta_{\text{Eval}}$ per candidate per
declaration in $\mathrm{d}_i$, which for Figure~\ref{fig:hyb_example} is five
candidates against two declarations. Answers are cached on the pair of declaration
and arguments, so a candidate appearing in the option lists of several neighbors is
queried once. Sports Understanding admits two outcomes and its skeletons contain no
such loop, issuing one call per declaration for the whole instance. The loop follows
from the number of candidates the task presents, and both benchmarks pass through the
same intervention and execution.

\section{More Experiment of Hybrid Program }
\label{app:instantiation}

This section details the hybrid abstraction of
Section~\ref{sec:instantiation}. Its settings are given in
Section~\ref{app:implementation_details}.

\subsection{Static Validation}
Three constraints are enforced by inspecting the abstract syntax tree rather than
by prompt instruction, and a candidate failing any of them after the retries of Appendix~\ref{app:implementation_details} is discarded. The
program body contains at least one deterministic construct. Every call to an
undefined function takes at most two arguments, each a single string or number, and
no argument is the question, the option list, or any name bound to either along an
assignment chain. Every undefined name called in the body is declared, carries
complete type annotations and a non-empty docstring, and returns \texttt{bool},
\texttt{int}, \texttt{float} or a short \texttt{str}. The second constraint is what
confines $\Theta_{\text{Eval}}$ to its arguments. Following the outcome-oriented
criterion of Section~\ref{sec:invariant_reasoning_distillation}, we additionally
require the abstraction to reproduce the answer on its own instance. The \textit{+ Code} condition of Table~\ref{tab:instantiation} uses the template of
Appendix~\ref{app:prompt_hybrid_abstraction} without the declaration block, so the
abstraction is a deterministic program, and candidates are retained by observational
correctness alone.

\subsection{Run-Time Evaluation}
$\Theta_{\text{Adapter}}$ returns the call line alone, and the declarations and the
body are reused verbatim. Since $\mathrm{s}_i$ returns the text of the selected
option rather than its label, we match the returned string against the option list of
$x_j$ and take the label at the matching position, so a return value absent from that
list yields $S_{\text{inv}} = 0$ in the same way as a failed execution.

\subsection{Diagnostics}
Observational filtering retains a similar share whether or not the demonstration
carries a program, 79.1\% without and 77.1\% with, so the reduction to 58.8\% under
ECI2D follows from the intervention rather than from the difficulty of producing the
abstraction. Table~\ref{tab:hyb_diag} decomposes the interventions that score zero.
Wrong options dominate and execution failures account for 5\% together, so what the
criterion rejects is a rationale that does not transfer rather than a program that
does not run. The retained skeletons combine their undefined functions over the
candidate answers rather than sequentially, with one call rarely informing the
argument of another, so what Step~2 transplants in this domain is the choice of
functions and the way their returned values are combined.

\begin{table}[h]
\centering
\caption{Decomposition of the interventions scoring $S_{\text{inv}} = 0$ under the
hybrid abstraction on CommonsenseQA.}
\label{tab:hyb_diag}
\small
\renewcommand{\arraystretch}{1.15}
\setlength{\tabcolsep}{4pt}
\begin{tabular}{@{}lc@{}}
\toprule
Cause & Share (\%) \\
\midrule
Wrong option returned & 94.92 \\
Execution timeout     & 3.57 \\
Code error            & 1.51 \\
\bottomrule
\end{tabular}
\end{table}

\subsection{Case Study}
Figure~\ref{fig:hyb_example} traces one retained candidate. The abstraction
declares two functions and scores the answer options by the values they return, and
the adapter rewrites only the call line. On the neighbor shown, the ground truth is
selected over a competing option that also scores, so the outcome follows from the
comparison rather than from a default.

A second candidate illustrates the separation of $\mathrm{s}_i$ from
$\mathrm{v}_i$. Its skeleton declares
\texttt{is\_pleasant\_person(person)} alongside two functions over the options, and
the adapter rebinds that argument from \texttt{receptionist lady} to \texttt{Jenny}
while leaving the body untouched. Its interventions also draw values from the cache,
since \texttt{office} appears among the options of several neighbors.

\begin{figure}[ht]
\centering
\begin{minipage}{\columnwidth}
\scriptsize

\noindent\textbf{Source instance ($x_i$, gold E)}
\begin{itemize}[leftmargin=1.2em, label=\textbullet, nosep, itemsep=2pt]
    \item \textbf{Question:} She had to comfort friend that day, she gave her a big ole what?
    \item \textbf{Options:} A) friendly \ B) sympathetic \ C) handshake \ D) cook \ E) hug
    \item \textbf{Abstraction ($\hat{p}_i$):}
    \begin{lstlisting}[gobble=4, basicstyle=\ttfamily\scriptsize]
    def is_physical_action(action: str) -> bool:
        """Returns True if the action involves physical contact."""
    def is_emotional_support(action: str) -> bool:
        """Returns True if the action provides emotional support."""

    def solution(param, options):
        best, best_score = options[0], -1.0
        for option in options:
            score = 0.0
            if is_physical_action(option):   score += 1.0
            if is_emotional_support(option): score += 1.0
            if score > best_score:
                best, best_score = option, score
        return best
    \end{lstlisting}
    \item \textbf{Self-execution:} returns \texttt{E) hug}, matching the gold answer.
\end{itemize}

\vspace{0.3em}
\hrule
\vspace{0.3em}

\noindent\textbf{Intervention on a neighbor ($x_j$, gold D)}
\begin{itemize}[leftmargin=1.2em, label=\textbullet, nosep, itemsep=2pt]
    \item \textbf{Question:} James and Lilly have strong feelings for each other. To express those feelings they kiss. What else might they do?
    \item \textbf{Adapted call ($\hat{p}_{i \to j}$):}
    \begin{lstlisting}[gobble=4, basicstyle=\ttfamily\scriptsize]
    solution("kiss", ["shake hands", "fall in love", "smile",
                      "embrace", "falling in love"])
    \end{lstlisting}
    \item \textbf{Returned values:} \texttt{shake hands} scores 1.0, \texttt{embrace} scores 2.0, the remaining options score 0.0.
    \item \textbf{Result:} returns \texttt{embrace}, matching $y_j$, so $S_{\text{inv}} = 1$.
\end{itemize}

\end{minipage}
\caption{One retained candidate under the hybrid abstraction. The logical skeleton
is transplanted unchanged and only the call line is rewritten.}
\label{fig:hyb_example}
\end{figure}

\section{More analysis of Observational Equivalence under Observation Distillation}
\label{app:derivation}
We formally analyze the observational equivalence of the causal structure governing the generation of rationales under the constraint of observational data. Let $\mathcal{D} = \{(x_i, r_i, y_i)\}_{i=1}^{|\mathcal{D}|}$ denote a dataset of demonstrations distilled via an outcome-oriented criterion, where $y_i$ matches the ground truth $y_{gt}$. We posit two competing structural causal models (SCMs) that could generate this data: the Faithful Causal Model $\mathcal{M}_{\text{causal}}$ and the Spurious Post-hoc Model $\mathcal{M}_{\text{spurious}}$. Under the Faithful Causal Model $\mathcal{M}_{\text{causal}}$, the rationale $R$ functions as a necessary mediator that transmits the causal influence of input $X$ to output $Y$. The joint probability distribution factorizes according to the directed acyclic graph $X \to R \to Y$:
\begin{equation}
    P_{\mathcal{M}_{\text{causal}}}(X, R, Y) = P(Y \mid R) P(R \mid X) P(X).
\label{eq:causal_factorization}
\end{equation}
This factorization implies a conditional independence structure where the output is independent of the input given the rationale, denoted as $(Y \perp X \mid R)$. Conversely, under the \textit{Spurious Post-hoc Model} $\mathcal{M}_{\text{spurious}}$, the output $Y$ is determined directly by the input $X$ (potentially via latent shortcuts), while the rationale $R$ is generated retrospectively to justify the observed $(X, Y)$ pair. This structure corresponds to the graph $X \to Y \to R$ or a confounding structure where $R$ is a collider $X \to R \leftarrow Y$. The joint probability distribution for the post-hoc chain $X \to Y \to R$ factorizes as:
\begin{equation}
P_{\mathcal{M}_{\text{spurious}}}(X, R, Y) = P(R \mid X, Y) P(Y \mid X) P(X).
\label{eq:spurious_factorization}
\end{equation}
The fundamental challenge of observation distillation arises from the fact that for any distribution $P(X, R, Y)$ generated by a faithful model, there exists a parameterization of a spurious model that can reproduce the exact same observational distribution. By the chain rule of probability, the joint distribution can generally be decomposed as $P(X, R, Y) = P(R \mid X, Y) P(Y \mid X) P(X)$. Without external constraints or interventional data, the faithful factorization in Equation~\ref{eq:causal_factorization} is mathematically indistinguishable from the spurious factorization in Equation~\ref{eq:spurious_factorization} solely based on the observed samples. This phenomenon corresponds to the concept of Markov Equivalence Classes in causal discovery~\cite{spirtes2000causation, verma2022equivalence}, where multiple causal graphs entail the same set of conditional independencies and thus generate identical observational likelihoods. Consequently, an outcome-oriented filter that selects demonstrations maximizing $P(Y = y_{gt} \mid X, R)$ cannot differentiate between a rationale that \textit{causes} the correct answer ($R \to Y$) and a rationale that is \textit{caused by} the correct answer ($Y \to R$). Although these two models are observationally equivalent, they are interventionally distinct~\cite{scm, peters2017elements}. Specifically, the interventional distributions differ significantly:
\begin{equation}
\begin{split}
        P_{\mathcal{M}_{\text{causal}}}(Y \mid do(R=r), X) &= P(Y \mid r), \\
        P_{\mathcal{M}_{\text{spurious}}}(Y \mid do(R=r), X) &= P(Y \mid X).
\end{split}
\end{equation}

In the faithful case, intervening on $R$ alters the probability of $Y$, whereas in the spurious case, the intervention leaves the distribution of $Y$ invariant (or governed solely by $X$). This derivation theoretically justifies our proposed ECI2D framework, which utilizes the $do$-operator to break the observational equivalence and identify $R_c$ through the invariance of the interventional distribution.

\section{More analysis of Transfer Invariance from Independent Causal Mechanisms}
\label{app:icm_proof}

In this section, we provide a formal justification for utilizing transfer invariance as a proxy for rationale faithfulness. Our derivation relies on the Principle of Independent Causal Mechanisms (ICM)~\cite{independent_causal_mechanisms}, which posits that the conditional distribution of an effect given its direct causes remains invariant to changes in the marginal distribution of the causes. We analyze the behavior of faithful and spurious rationales under causal intervention across local distributional shifts.

Let $S$ denote a shift variable representing the transition from a source context $x_i$ to a neighbor context $x_j \in \mathcal{N}(x_i)$. This shift induces a change in the input distribution $P(X \mid S)$. We examine the stability of the conditional probability $P(Y \mid do(R), X)$ under this shift for two distinct causal structures.

\paragraph{Rationale as an Operational Procedure.}
Before the case analysis, we make explicit the operational sense in which a rationale acts as a causal mechanism. We treat a rationale $r \in \mathcal{R}$ as a function $r : \mathcal{X}_{\mathcal{C}} \to \mathcal{Y}_{\mathcal{C}}$ that maps inputs within a problem class $\mathcal{C}$, defined by shared logical features $A=(P, S, T)$, to their predicted outputs. Under this view, faithfulness with respect to $(x_i, y_i)$ requires $r(x_j) = y_j$ for every $(x_j, y_j) \in \mathcal{C}$. A rationale that satisfies this requirement only on its source input provides no procedural content and reduces to a constant function returning $y_i$, which carries no explanatory role. Equation~\ref{eq:intervention} formalizes this requirement in probabilistic form by reducing it to the necessary condition that interventions with $r$ on neighbors $x_j \in \mathcal{N}(x_i)$ preserve correctness in distribution.

\paragraph{Case 1: Faithfulness Implies Invariance.}
Consider the faithful causal rationale $R_c$ defined in Definition~\ref{def:faithful}, where the causal structure is governed by the directed edge $R_c \to Y$. According to the ICM principle, the causal mechanism $P(Y \mid R_c)$ is autonomous and modular, meaning it remains invariant to interventions or shifts in the input generation process $P(X)$.
Formally, when we perform an intervention $do(R=r)$ by transplanting a rationale skeleton extracted from $x_i$ to $x_j$, the interventional distribution factorizes as:
\begin{equation}
    P(Y \mid do(R_c=r), x_j) = P(Y \mid r).
\end{equation}
Since the mechanism $P(Y \mid R)$ captures the underlying functional law (e.g., the arithmetic logic of addition) rather than statistical correlations, it remains valid regardless of the specific parameterization of the input $x_j$. Consequently, a faithful rationale demonstrates transfer invariance, as the intervened mechanism consistently induces the correct ground truth $y_j$ across the local neighborhood $\mathcal{N}(x_i)$.

\paragraph{Case 2: Spuriousness Implies Variance.}
Conversely, consider the spurious post-hoc rationale $R_s$ defined in Definition~\ref{def:spurious}, which arises from the reverse causal path $Y \to R_s$ or confounding by $X$. In this scenario, the rationale is functionally dependent on the specific outcome $y_i$ of the source context.
Applying the $do$-operator to set $R_s = r$ breaks the incoming edges to $R$, rendering $R$ statistically independent of its original parents $X$ and $Y$ in the interventional graph. The distribution of $Y$ is thus governed solely by the input $X$ via the spurious mechanism $\mathcal{M}_{\text{spurious}}$:
\begin{equation}
    P(Y \mid do(R_s=r), x_j) = P(Y \mid x_j).
\end{equation}
However, the explicit content of the intervened rationale $r$ carries the specific context of $x_i$. Since $R_s$ does not structurally cause $Y$ but merely correlates with it in the source distribution, forcing this rationale upon a new context $x_j$ creates a mismatch between the logic encoded in $r$ and the ground truth $y_j$. The observational correlation $P(Y \mid R)$ observed at $x_i$ fails to transport to $x_j$ because the underlying joint distribution $P(X, Y, R)$ changes with $S$. Therefore, spurious rationales violate transfer invariance, exhibiting a performance degradation when subjected to causal intervention.

\paragraph{Sound but Incomplete Identification.}
The two cases above establish the asymmetric identifiability of $R_c$ and $R_s$ under transfer invariance. Let $T(\hat{r}_i) \in \{0, 1\}$ denote the invariance verdict, where $T(\hat{r}_i) = 1$ if $\hat{r}_i$ satisfies Equation~\ref{eq:intervention} across $\mathcal{N}(x_i)$. Case 1 establishes the implication $\hat{r}_i = R_c \Rightarrow T(\hat{r}_i) = 1$, whose contrapositive $T(\hat{r}_i) = 0 \Rightarrow \hat{r}_i \neq R_c$ provides a soundness guarantee for the filter and does not depend on the absence of latent variables. Case 2 establishes that a generic post-hoc $R_s$ violates Equation~\ref{eq:intervention}, yet does not preclude the existence of pathological rationales that share statistical regularities with $R_c$ across $\mathcal{N}(x_i)$ without being causally faithful, for instance through unobserved factors that act uniformly across the neighborhood. Equation~\ref{eq:intervention} therefore provides a necessary but not sufficient condition for faithfulness, and ECI2D should be understood as a sound filter that excludes the most prevalent non-faithful rationales rather than a complete identifier of the causal mechanism. The residual configurations are discussed in Limitations.

\paragraph{Empirical Estimation in Finite Samples.}
Direct computation of the probabilities in Equation~\ref{eq:intervention} is infeasible for an LLM-induced distribution, so Section~\ref{sec:method} replaces them with the empirical estimator
\begin{equation}
\hat{S}_N(\hat{r}_i) = \frac{1}{N} \sum_{j=1}^{N} \mathbb{I}\!\left(\hat{y}_j = y_j\right),
\end{equation}
where $\{x_j\}_{j=1}^N$ are drawn from $\mathcal{N}(x_i)$ and $\hat{y}_j = \mathrm{Exec}(\hat{p}_{i \to j})$. Each indicator is an unbiased single-sample estimator of $P(Y=y_j \mid do(R=\hat{r}_i), X=x_j)$ at neighbor $x_j$. At the small values of $N$ used in practice the estimator is coarse, and we therefore adopt the strictest available threshold, requiring agreement on every neighbor except on DROP, where the F1 metric admits a threshold of $0.9$. Sensitivity to $N$ is reported in Section~\ref{sec:results}.

\section{Computational Cost Analysis}
\label{app:cost}
The ECI2D framework incurs additional computational overhead strictly during the offline demonstration distillation phase. This overhead originates from iterative rationale sampling, program abstraction and adaptation, logical neighbor retrieval, and repeated sandbox executions for verification. For a standard benchmark such as GSM8K with a logical neighborhood size of $N=3$, the pipeline entails approximately 22,000--30,000 lightweight Python operations. Execution of these scripts requires less than one minute on a standard CPU, with the dominant cost arising from offline LLM generation calls.

Crucially, this preprocessing cost is strictly a one-time investment. During the online inference phase, ECI2D introduces no additional computational overhead compared to standard self-generated ICL. The model receives a standard few-shot prompt format, and no external execution or intervention steps are performed at test time. 

The offline computational investment yields a compact, causally validated demonstration pool. This distilled subset reduces online vector retrieval latency and decreases context token consumption. Consequently, the framework implements a standard offline-online trade-off: increased preprocessing cost is exchanged for higher demonstration fidelity and downstream reasoning robustness, while preserving the computational profile of conventional ICL during deployment.

\section{Human Validation}
\label{app:human}

All annotation was carried out by three annotators with graduate-level training in NLP, working independently and blind to the pipeline's output. 

\paragraph{Validity of neighborhood construction.} The construction of $\mathcal{N}(x_i)$ rests on the assumption that the two-stage retrieval identifies instances sharing a reasoning structure. We sampled $1{,}000$ pairs judged logically similar by the pipeline from each of GSM8K and DROP, and asked three annotators to decide independently, and blind to the pipeline's verdict, whether the two instances admit a common logical skeleton under $A=(P,S,T)$. Mean agreement with the pipeline is $94.4\%$ on GSM8K and $91.3\%$ on DROP (Table~\ref{tab:human}). Disagreements concentrate on instances with strong lexical overlap but divergent premise structure, which the semantic retriever ranks highly and the structural verifier does not always reject. The residual error rate propagates conservatively: a faithful rationale evaluated against a mismatched neighbor fails the execution check and is discarded, which lowers retention without admitting spurious rationales into $D_{\text{causal}}$.

\paragraph{Alignment between rationale and program.} A second assumption is that $\Pi$ formalizes the reasoning structure of $\hat r_i$ rather than altering it. Three annotators judged $2{,}000$ $\langle \hat r, \hat p \rangle$ pairs for step-level correspondence, reporting a mean alignment of $95.5\%$. Since $\hat r$ and $\hat p$ are emitted in a single forward pass (Appendix~\ref{app:prompt_gen_abs}), they are conditioned on the same context, which we expect accounts for the high agreement. Observed misalignments are of two kinds with opposite consequences. Alteration of the skeleton causes the adapted program to fail on its neighbors, so the demonstration is discarded. Silent repair, in which $\Pi$ maps a flawed $\hat r$ to a correct $\hat p$, has the opposite effect and is not detectable by alignment annotation, since the annotated program is correct by construction. We additionally inspected $500$ demonstrations sampled from $D_{\text{causal}}$ for chain-level correctness and observed no incorrect reasoning chain, which bounds the prevalence of silent repair in the retained pool at $0.76\%$ with $95\%$ confidence, but a direct measurement of this rate remains future work.

\begin{table}[t]
\centering
\caption{Human validation of the assumptions underlying ECI2D. Three annotators judged every sample independently and blind to condition. Ranges give the per-annotator minimum and maximum.}
\label{tab:human}
\small
\renewcommand{\arraystretch}{1.2}
\setlength{\tabcolsep}{3pt}
\begin{tabular}{@{}p{2.1cm}ccc@{}}
\toprule
Target & Source & Samples & Agreement \\
\midrule
\multirow{2}{2.1cm}{neighborhood membership}
 & GSM8K & 1{,}000 & 94.4 \small(93.6--95.3) \\
 & DROP  & 1{,}000 & 91.3 \small(90.3--91.8) \\
\midrule
$\langle \hat r,\hat p\rangle$ alignment & both & 2{,}000 & 95.5 \small(94.5--96.5) \\
\midrule
Chain correctness in $D_{\text{causal}}$ & GSM8K & 500 & 100.0 \small[99.24, 100] \\
\bottomrule
\end{tabular}
\end{table}

\paragraph{Case study of adapter errors.} Figure~\ref{fig:case_study} illustrates the two failure modes discussed above with instances drawn from GSM8K. In Case 1 the Adapter alters the arithmetic structure of the source skeleton, so the adapted program returns a value inconsistent with the ground truth and the source demonstration is discarded. In Case 2 the skeleton is preserved but the target instance expresses its parameter in a different unit, which the Adapter fails to normalize before substitution. Both are conservative in effect, since a failed execution yields $S_{\text{inv}} = 0$ and removes the candidate from $D_{\text{causal}}$.

\begin{figure}[ht]
\centering
% 使用 minipage 限制在单栏宽度内
\begin{minipage}{\columnwidth}

% --- 全局字体调小 ---
\scriptsize 

\noindent\textbf{Case 1: Skeleton Alteration}
\begin{itemize}[leftmargin=1.2em, label=\textbullet, nosep, itemsep=2pt]
    \item \textbf{Source Problem ($x_i$):} Janet's ducks lay 16 eggs/day. She eats 3, uses 4 for baking, and sells the rest for \$2 each. Daily earnings?
    
    \item \textbf{Source Code ($\hat{p}_i$):}
    % gobble=4 用于消除左侧缩进，basicstyle 设置代码字体为等宽且极小
    \begin{lstlisting}[gobble=4, basicstyle=\ttfamily\scriptsize]
    def solution():
        laid, eaten, baked, price = 16, 3, 4, 2
        remainder = laid - eaten - baked
        return remainder * price
        # Logic: (a - b - c) * d
    \end{lstlisting}
    
    \item \textbf{Logically Similar Target ($x_j$):} A farm produces 40 gallons of milk/day. The farmer uses 5 for his family, gives 10 to calves, and sells the rest for \$3/gallon. Daily earnings?
    
    \item \textbf{Adapter Code ($\hat{p}_{i \to j}$):}
    \begin{lstlisting}[gobble=4, basicstyle=\ttfamily\scriptsize]
    def solution():
        produced, family, calves, price = 40, 5, 10, 3
        return produced - (family * price) - (calves * price)
        # Logic: (a * b) - (c * d)
    \end{lstlisting}
    
    \item \textbf{Error:} \textit{Adapter alters the causal skeleton and distributes the price incorrectly.}
    \item \textbf{Result:} Outputs -5 instead of the correct answer 75.
\end{itemize}

\vspace{0.3em}
\hrule
\vspace{0.3em}

\noindent\textbf{Case 2: Variable Mapping Failure}
\begin{itemize}[leftmargin=1.2em, label=\textbullet, nosep, itemsep=2pt]
    \item \textbf{Source Problem ($x_i$):} Weng earns \$12/hour. She worked for 50 minutes. How much did she earn?
    
    \item \textbf{Source Code ($\hat{p}_i$):}
    \begin{lstlisting}[gobble=4, basicstyle=\ttfamily\scriptsize]
    def solution():
        rate, minutes = 12, 50
        return rate * (minutes / 60)
    \end{lstlisting}
    
    \item \textbf{Logically Similar Target ($x_j$):} A worker is paid \$15/hour. He worked for 3 hours and 20 minutes. How much was he paid?
    
    \item \textbf{Adapter Code ($\hat{p}_{i \to j}$):}
    \begin{lstlisting}[gobble=4, basicstyle=\ttfamily\scriptsize]
    def solution():
        rate = 15
        minutes = 20
        return rate * (minutes / 60)
    \end{lstlisting}
    
    \item \textbf{Error:} \textit{Adapter fails to synthesize "3 hours and 20 mins" into a single minute variable to fit the skeleton.}
    \item \textbf{Result:} Outputs 5 instead of the correct answer 50.
\end{itemize}

\end{minipage}
\caption{Case study of LLM adapter errors.}
\label{fig:case_study}
\end{figure}

\newpage

\section{Prompts}
\label{app:prompts}

\subsection{Rationales Generation and Abstraction}
\label{app:prompt_gen_abs}

\begin{tcolorbox}[
    colback=white,             % 内容背景色：白色
    colframe=prompt-gray,      % 边框和标题背景色：深灰
    coltitle=white,            % 标题文字颜色：白色
    title=\textbf{Prompt template of Rationales Generation and Abstraction for Math Reasoning}, % 标题内容
    fonttitle=\large,          % 标题字体大小
    boxrule=1.5pt,             % 边框粗细
    arc=3pt,                   % 圆角大小
    enhanced,                  % 启用高级绘图引擎
    fontupper=\ttfamily\small, % 全局内容字体：打字机字体 (Typewriter)，字号偏小
    left=5mm, right=5mm, top=5mm, bottom=5mm, % 内边距
]

\textbf{\rmfamily Instruction} \\
You are a Python Expert. Write a generic Python function `def solution():` to solve this math problem. \\
First, think step-by-step about the logic required (Chain of Thought). \\
Then, write a Python function `def solution():` that returns the answer. \\
Question: [question] \\
Format your response as: \\
Rationale: \\
... content ... \\
Code: \\
```python \\
... code ... \\
```
\end{tcolorbox}

\begin{tcolorbox}[
    colback=white,             % 内容背景色：白色
    colframe=prompt-gray,      % 边框和标题背景色：深灰
    coltitle=white,            % 标题文字颜色：白色
    title=\textbf{Prompt template of Rationales Generation and Abstraction for Logic Reasoning.}, % 标题内容
    fonttitle=\large,          % 标题字体大小
    boxrule=1.5pt,             % 边框粗细
    arc=3pt,                   % 圆角大小
    enhanced,                  % 启用高级绘图引擎
    fontupper=\ttfamily\small, % 全局内容字体：打字机字体 (Typewriter)，字号偏小
    left=5mm, right=5mm, top=5mm, bottom=5mm, % 内边距
]

\textbf{\rmfamily Instruction} \\
You are a Python Expert. Solve the problem based on the Context. \\
First, think step-by-step about the logic required (Chain of Thought). \\
Then, write a Python function `def solution():` that returns the answer. \\
Use `datetime`, `math`, `sorting`, or `counting` logic as needed. \\
Context: \{passage\} \\
Question: \{question\} \\
Format your response as: \\
CoT: \\
... content ... \\
Code: \\
```python \\
... code ... \\
```
\end{tcolorbox}

\newpage
\subsection{Adaptation and Intervention}
\label{app:prompt_adapt}

\begin{tcolorbox}[
    colback=white,             % 内容背景色：白色
    colframe=prompt-gray,      % 边框和标题背景色：深灰
    coltitle=white,            % 标题文字颜色：白色
    title=\textbf{Prompt template of Adaptation and Intervention}, % 标题内容
    fonttitle=\large,          % 标题字体大小
    boxrule=1.5pt,             % 边框粗细
    arc=3pt,                   % 圆角大小
    enhanced,                  % 启用高级绘图引擎
    fontupper=\ttfamily\small, % 全局内容字体：打字机字体 (Typewriter)，字号偏小
    left=5mm, right=5mm, top=5mm, bottom=5mm, % 内边距
]

\textbf{\rmfamily Instruction} \\
Solve the NEW PROBLEM using the LOGIC of the EXISTING CODE. \\
1. Analyze the NEW PROBLEM. \\
2. Reuse the structure of EXISTING CODE (do not change the formulas/logic).  \\
3. Call the function with values from the NEW PROBLEM. \\
EXISTING CODE: \\
```python \\
\{retrieved\_code\} \\
``` \\
NEW PROBLEM: \\
\{retrieved\_question\} \\
Output Python Code based on the Existing Code's logic: \\
```python \\
code \\
```
\end{tcolorbox}

\newpage

\subsection{Causal Steering Inference}
\label{app:prompt_reasoning}

\begin{tcolorbox}[
    colback=white,             % 内容背景色：白色
    colframe=prompt-gray,      % 边框和标题背景色：深灰
    coltitle=white,            % 标题文字颜色：白色
    title=\textbf{Prompt template of Math Reasoning}, % 标题内容
    fonttitle=\large,          % 标题字体大小
    boxrule=1.5pt,             % 边框粗细
    arc=3pt,                   % 圆角大小
    enhanced,                  % 启用高级绘图引擎
    fontupper=\ttfamily\small, % 全局内容字体：打字机字体 (Typewriter)，字号偏小
    left=5mm, right=5mm, top=5mm, bottom=5mm
]

% 1. Instruction 部分
\textbf{\rmfamily Instruction} \\ % 使用 \rmfamily 让标题变为非等宽字体，更显眼
You are a Math and Reasoning Expert utilizing a Hybrid Reasoning approach. \\
Combine natural language intuition with Python precision. \\
Reference the provided examples for reasoning style. \\
You should:  \\
Natural Language Analysis: Explain the reasoning step-by-step (chain-of-thought). \\
Python Implementation: Consider the logic in code. \\
CRITICAL: Do NOT output python code to execute.  \\
At the very end of your response, strictly output the answer in this format: \\
Final Answer:

\vspace{0.4cm} % 增加段落间距

% 2. Demonstration 部分
\textbf{\rmfamily Demonstration} \\
--- Example i+1 --- \\
Question: \{question\} \\
CoT: \{rationale\} \\
Code: \{code\} \\

\vspace{0.4cm} % 增加段落间距

--- Current Task --- \\ 
Question: \{question\} \\
Think step by step, ensure precision. \\
Format: Final Answer:

\end{tcolorbox}

\begin{tcolorbox}[
    colback=white,             % 内容背景色：白色
    colframe=prompt-gray,      % 边框和标题背景色：深灰
    coltitle=white,            % 标题文字颜色：白色
    title=\textbf{Prompt template of DROP Reasoning}, % 标题内容
    fonttitle=\large,          % 标题字体大小
    boxrule=1.5pt,             % 边框粗细
    arc=3pt,                   % 圆角大小
    enhanced,                  % 启用高级绘图引擎
    fontupper=\ttfamily\small, % 全局内容字体：打字机字体 (Typewriter)，字号偏小
    left=5mm, right=5mm, top=5mm, bottom=5mm
]

% 1. Instruction 部分
\textbf{\rmfamily Instruction} \\ % 使用 \rmfamily 让标题变为非等宽字体，更显眼
You are a Logic and Reasoning Expert.  \\
Solve the reading comprehension problem based on the Context. \\
Reference the provided examples for reasoning style. \\
Do NOT output python code to execute. \\
At the very end of your response, strictly output the answer in this format: \\
Final Answer:

\vspace{0.4cm} % 增加段落间距

% 2. Demonstration 部分
\textbf{\rmfamily Demonstration} \\
--- Example i+1 --- \\
Context: \{passage\} \\
Question: \{question\} \\
CoT: \{rationale\} \\
Code: \{code\} \\

\vspace{0.4cm} % 增加段落间距

--- Current Task --- \\ 
Context: \{passage\} \\
Question: \{question\} \\
Think step by step, ensure precision. \\
Format: Final Answer:

\end{tcolorbox}

\begin{tcolorbox}[
    colback=white,             % 内容背景色：白色
    colframe=prompt-gray,      % 边框和标题背景色：深灰
    coltitle=white,            % 标题文字颜色：白色
    title=\textbf{Prompt template of BBH logic subset Reasoning}, % 标题内容
    fonttitle=\large,          % 标题字体大小
    boxrule=1.5pt,             % 边框粗细
    arc=3pt,                   % 圆角大小
    enhanced,                  % 启用高级绘图引擎
    fontupper=\ttfamily\small, % 全局内容字体：打字机字体 (Typewriter)，字号偏小
    left=5mm, right=5mm, top=5mm, bottom=5mm % 内边距
    % breakable                  % 允许跨页 (如果附录很长，这很重要)
]

\textbf{\rmfamily Instruction} \\ 
You are an expert in logical reasoning and causal inference. \\
I will provide you with REFERENCE EXAMPLES that include a Question, Analysis (CoT), and Pseudo-Code. \\
NOTE: The reference examples might be from a different domain (e.g., reading comprehension), \\
but you must abstract the **reasoning logic** and **algorithmic thinking** from them. \\
Apply this logic to solve the CURRENT TASK step-by-step. \\
Do NOT execute python code. Use the code only as a logical blueprint. \\

\vspace{0.4cm} % 增加段落间距

% 2. Demonstration 部分
\textbf{\rmfamily Demonstration} \\
--- Example i+1 --- \\
Context: \{passage\} \\
Question: \{question\} \\
CoT: \{rationale\} \\
Code: \{code\} \\
Note: Learn the reasoning logic (how to calculate/count/sort), even if the context format is different.\\

\vspace{0.4cm} % 增加段落间距

--- Current Task --- \\ 
Context: \{passage\} \\
Question: \{question\} \\
Think step by step, ensure precision. \\
Format: Final Answer:

\end{tcolorbox}

\begin{tcolorbox}[
    colback=white,
    colframe=prompt-gray,
    coltitle=white,
    title=\textbf{Prompt template of Commonsense and BBH semantic subset Reasoning},
    fonttitle=\large,
    boxrule=1.5pt,
    arc=3pt,
    enhanced,
    fontupper=\ttfamily\small,
    left=5mm, right=5mm, top=5mm, bottom=5mm,
]

\textbf{\rmfamily Instruction} \\
You are a Commonsense Reasoning Expert utilizing a Hybrid Reasoning approach. \\
Combine natural language intuition with Python precision. \\
Reference the provided examples for reasoning style. \\
The helper functions in the examples are not implemented, so treat them as judgements you make yourself. \\
Do NOT output python code to execute. \\
At the very end of your response, strictly output the answer in this format: \\
Final Answer:

\vspace{0.4cm}

\textbf{\rmfamily Demonstration} \\
--- Example i+1 --- \\
Question: \{question\} \\
Options: \{options\} \\
CoT: \{rationale\} \\
Code: \{primitives\} \{code\} \\

\vspace{0.4cm}

--- Current Task --- \\
Question: \{question\} \\
Options: \{options\} \\
Think step by step, ensure precision. \\
Format: Final Answer:
\end{tcolorbox}

\newpage

\subsection{Hybrid Abstraction}
\label{app:prompt_hybrid_abstraction}

\begin{tcolorbox}[
    colback=white,
    colframe=prompt-gray,
    coltitle=white,
    title=\textbf{Prompt template of Rationales Generation and Hybrid Abstraction},
    fonttitle=\large,
    boxrule=1.5pt,
    arc=3pt,
    enhanced,
    fontupper=\ttfamily\small,
    left=5mm, right=5mm, top=5mm, bottom=5mm,
]

\textbf{\rmfamily Instruction} \\
You are a Python Expert. Solve the multiple-choice question by writing a generic Python function `def solution():`. \\
First, think step-by-step about the logic required (Chain of Thought). \\
Then, declare any step that requires world knowledge rather than computation as an undefined helper function, with a type signature and a docstring. \\
Finally, write `def solution():` that scores each option using those helpers and returns the highest scoring one. \\
Rules: \\
1. Each helper takes at most two arguments, each a single string or number. \\
2. Never pass the question or the full option list to a helper. \\
3. Accumulate scores as floats and compare with `>`, so ties keep the first option. \\
4. Always return an option, never `None`. \\
Question: \{question\} \\
Options: \{options\} \\
Format your response as: \\
Rationale: \\
... content ... \\
Primitives: \\
```python \\
... declarations only ... \\
``` \\
Code: \\
```python \\
... def solution ... \\
``` \\
Call: \\
```python \\
... solution(...) ... \\
```
\end{tcolorbox}

\subsection{Adaptation and Intervention for Hybrid Programs}
\label{app:prompt_hybrid_intervention}

\begin{tcolorbox}[
    colback=white,
    colframe=prompt-gray,
    coltitle=white,
    title=\textbf{Prompt template of Adaptation and Intervention for Hybrid Programs},
    fonttitle=\large,
    boxrule=1.5pt,
    arc=3pt,
    enhanced,
    fontupper=\ttfamily\small,
    left=5mm, right=5mm, top=5mm, bottom=5mm,
]

\textbf{\rmfamily Instruction} \\
Bind the NEW PROBLEM to the EXISTING CODE. \\
1. Analyze the NEW PROBLEM. \\
2. Do NOT change the helper declarations or the function body. \\
3. Output only the call line, with values from the NEW PROBLEM. \\
EXISTING CODE: \\
```python \\
\{primitives\} \\
\{code\} \\
``` \\
NEW PROBLEM: \\
\{retrieved\_question\} \\
Options: \{retrieved\_options\} \\
Output the call line only: \\
```python \\
solution(...) \\
```
\end{tcolorbox}

\subsection{Run-Time Function Evaluation}
\label{app:prompt_fun_eval}

\begin{tcolorbox}[
    colback=white,
    colframe=prompt-gray,
    coltitle=white,
    title=\textbf{Prompt template of Run-Time Function Evaluation},
    fonttitle=\large,
    boxrule=1.5pt,
    arc=3pt,
    enhanced,
    fontupper=\ttfamily\small,
    left=5mm, right=5mm, top=5mm, bottom=5mm,
]

\textbf{\rmfamily Instruction} \\
Answer the following function call. \\
You are given the declaration of the function and its arguments, and nothing else. \\
Return a single value that matches the declared return type. \\
Do not explain. \\
Declaration: \\
```python \\
\{declaration\} \\
``` \\
Arguments: \{arguments\} \\
Return value:
\end{tcolorbox}

\newpage
\subsection{Similar Sample Fine-grained Retrieval}
\label{app:prompt_retriveal}
\begin{tcolorbox}[
    colback=white,             % 内容背景色：白色
    colframe=prompt-gray,      % 边框和标题背景色：深灰
    coltitle=white,            % 标题文字颜色：白色
    title=\textbf{Prompt Template for Logically Similar Sample Retrieval}, % 标题内容
    fonttitle=\large,          % 标题字体大小
    boxrule=1.5pt,             % 边框粗细
    arc=3pt,                   % 圆角大小
    enhanced,                  % 启用高级绘图引擎
    fontupper=\ttfamily\small, % 全局内容字体：打字机字体
    left=5mm, right=5mm, top=5mm, bottom=5mm, % 内边距
]

\textbf{Instruction} \\
You are a structural reasoning evaluator. Assess logical similarity between a Target Question and each candidate based on three formal components:
\begin{enumerate}[leftmargin=*, nosep, topsep=2pt]
    \item \textbf{Premises ($P$):} Initial conditions, constraints, or given facts.
    \item \textbf{Subjects ($S$):} Core entities or variables under analysis.
    \item \textbf{Result Type ($T$):} Required output format or objective.
\end{enumerate}
Two questions are logically similar ($\theta=0$) if their $(P, S, T)$ structures are equivalent. Ignore surface phrasing, domain context, or specific numerical values. Compute $\theta \in \{0, 1\}$ for each candidate. Return the indices of the top-$N$ candidates where $\theta=0$ as a Python list.
\vspace{0.4cm} % 增加段落间距
Target Question: \{question\} \\
Related Questions: \{related\_questions\_List\} \\
Output:
\end{tcolorbox}

\end{document}